\documentclass{article}

     \usepackage[preprint,nonatbib]{neurips_2020}

\usepackage[numbers]{natbib}
\usepackage[utf8]{inputenc} % allow utf-8 input
\usepackage[T1]{fontenc}    % use 8-bit T1 fonts
\usepackage{hyperref}       % hyperlinks
\usepackage{url}            % simple URL typesetting
\usepackage{booktabs}       % professional-quality tables
\usepackage{amsfonts}       % blackboard math symbols

\usepackage{microtype}      % microtypography
\usepackage{graphicx}
\usepackage{nicefrac}       % compact symbols for 1/2, etc.
\usepackage{amsthm}
\usepackage{amsmath}
\usepackage{amssymb}
\usepackage{bbm}
\usepackage{wrapfig}

\usepackage{float}        %added by Robin
\usepackage{subcaption}     %added by Robin
\usepackage{graphicx}      %added by Robin
\usepackage{dcolumn}% Align table columns on decimal point
\usepackage{bm}% bold math
\usepackage{color}

\usepackage{algorithm}
\usepackage{algorithmic}
\usepackage{footnote}

\usepackage{amsmath,amsfonts,bm}

\def\eqref#1{equation~\ref{#1}}
\def\1{\bm{1}}

\def\vg{{\bm{g}}}

\def\vv{{\bm{v}}}
\def\vw{{\bm{w}}}
\def\vx{{\bm{x}}}
\def\vy{{\bm{y}}}
\def\vz{{\bm{z}}}
\def\vphi{{\bm{\phi}}}

\def\mH{{\bm{H}}}

\def\mM{{\bm{M}}}

\DeclareMathAlphabet{\mathsfit}{\encodingdefault}{\sfdefault}{m}{sl}
\SetMathAlphabet{\mathsfit}{bold}{\encodingdefault}{\sfdefault}{bx}{n}

\DeclareMathOperator{\Tr}{Tr}

\newtheorem{theorem}{Theorem}

\theoremstyle{remark}

{
	\theoremstyle{plain}
	
}

\title{Flatness is a False Friend}

\author{%
  Diego Granziol\thanks{\href{http://www.granziol.com}{Personal Homepage}} \\
Machine Learning Research Group\\
Oxford\\
United Kingdom \\
\texttt{diego@robots.ox.ac.uk} \\
}

\begin{document}

\maketitle
\begin{abstract}
%   The abstract paragraph should be indented \nicefrac{1}{2}~inch (3~picas) on
%   both the left- and right-hand margins. Use 10~point type, with a vertical
%   spacing (leading) of 11~points.  The word \textbf{Abstract} must be centered,
%   bold, and in point size 12. Two line spaces precede the abstract. The abstract
%   must be limited to one paragraph.
Hessian based measures of flatness, such as the trace, Frobenius and spectral norms, have been argued, used and shown to relate to generalisation. In this paper we demonstrate that for feed forward neural networks under the cross entropy loss, we would expect low loss solutions with large weights to have small Hessian based measures of flatness. This implies that solutions obtained using $L2$ regularisation should in principle be sharper than those without, despite generalising better. We show this to be true for logistic regression, multi-layer perceptrons, simple convolutional, pre-activated and wide residual networks on the MNIST and CIFAR-$100$ datasets. Furthermore, we show that for adaptive optimisation algorithms using iterate averaging, on the VGG-$16$ network and CIFAR-$100$ dataset, achieve superior generalisation to SGD but are $30 \times$ sharper. This theoretical finding, along with experimental results, raises serious questions about the validity of Hessian based sharpness measures in the discussion of generalisation. We further show that the Hessian rank can be bounded by the a constant times number of neurons multiplied by the number of classes, which in practice is often a small fraction of the network parameters. This explains the curious observation that many Hessian eigenvalues are either zero or very near zero which has been reported in the literature.
\end{abstract}
\section{Introduction}
\label{introduction}

Deep Neural Networks, despite parameter counts far exceeding the number of data-points and the fact they are trained with many passes of the same data, manage to post exceptional performance on held out test data. Quite why and how they generalise so well, remains an open question \citep{neyshabur2017exploring}. However, DNNs are not immune to the classical problem of over-fitting. \citet{zhang2016understanding} show that DNNs can perfectly fit random labels. Schedules with low initial and sharply decaying learning rates, lead to identical training but much higher testing error \citep{berrada2018deep,granziol2020towards,jastrzebski2020the}. In \citet{wilson2017marginal} the authors argue that specific adaptive gradient optimisers lead to solutions which don't generalise. This has lead to a significant development in partially adaptive algorithms \citep{chen2018closing,keskar2017improving}. 

Given the importance of accurate predictions on unseen data, understanding exactly what helps deep networks generalise has been a fundamental area of research. A key concept which has taken a foothold in the community, allowing for the comparison of different training loss minima using only the training data, is the concept of \textit{flatness}. From both a Bayesian and minimum description length framework, flatter minima generalize better than sharp minima \citep{hochreiter1997flat}.
The connection between flatness and generalisation is a key motivation behind many many optimization algorithm design choices, including both Entropy-SGD \citep{chaudhari2016entropy} and the use of Polyak averaging \citep{izmailov2018averaging}. \citet{keskar2016large} consider how large batch vs small batch stochastic gradient descent (SGD) alters the sharpness of solutions, with smaller batches leading to convergence to flatter solutions, leading to better generalization. \citet{jastrzkebski2018relation} look at the importance of the ratio learning rate and batch size in terms of generalization, finding that large ratios lead to flatter minima and better generalization. \citet{wu2017towards} consider the logarithm of the product of the top $k$ eigenvalues as a proxy measure for its volume (a truncated log determinant). \citet{yao2018hessian} investigated flat regions of weight space showing them to be more robust under adversarial attack. \citet{zhang2018theory} show that SGD concentrates in probability on flat minima. Mathematically when integrating out the product of the maximum likelihood (MLE) solution (given by the final weights) with the prior, the posterior is \textit{shifted} relative to the MLE solution. For \textit{sharp} minima, the difference in loss even for a small shift is potentially very large, where the sharpness is usually measured by properties of the second derivative of the loss, known as the \textit{Hessian}, such as the spectral norm or trace. The idea of a \textit{shift} between the training and testing loss surface is prolific in the literature and regularly related to generalisation \citep{he2019asymmetric,izmailov2018averaging,maddox2019simple}.

\paragraph{Contributions:} In this paper, we highlight that Hessian re-parameterisation arguments, hold in equal effect for the gradient.
Any gradient step can be arbitrarily altered in direction and magnitude with re-parameterisation.
Since practitioners extensively adopt gradient descent to 
find good model solutions, this does not constitute a strong argument against the practical use of Hessian based sharpness measures for generalisation metrics. Furthermore, given that in practice parameterisation is fixed, it is unclear whether this re-parameterisation has any effect in practice.
The question we seek to answer, is \emph{are Hessian based sharpness metrics at the end of training meaningful metrics for generalisation?} In this paper
\begin{itemize}
    \item We show that for the feed forward neural network with cross entropy loss we would expect Hessian based sharpness measures (such as the trace, Frobenius and spectral norms) to be reduced as the weights increase in magnitude and the loss is driven to $0$
    \item Based on this insight, we demonstrate that $L2$ regularisation, known to increase generalisation \citep{krogh1992simple}, increases Hessian based sharpness metrics, whilst also increasing generalisation for logistic regression, feed forward neural networks, simple convolutional networks, pre-activated and wide residual networks.
    \item For networks that employ batch normalisation, we show that when changing batch norm training mode to evaluation mode for curvature evaluation, that the extent of this effect is massively magnified
    \item We show that novel optimisation algorithms, that combine adaptive algorithms with iterate averaging \citet{granziol2020iterate} to achieve superior generalisation, can give $30\times$ greater Hessian based sharpness metrics and still generalise better. This effect is also more extreme in evaluation batch normalisation mode.
\end{itemize}

\paragraph{Related work:} The Hessians lack of reparameterisation invariance \citep{dinh2017sharp}, has subjected its use for predicting generalisation to criticism \citep{neyshabur2017exploring,tsuzuku2019normalized,rangamani2019scale}. Normalized definitions of flatness, have been introduced \citet{tsuzuku2019normalized,rangamani2019scale} in a PAC-Bayesian framework, although \citet{rangamani2019scale} note that empirically Hessian based sharpness measures correlate with generalisation. \citet{smith2017bayesian} argue that although sharpness can be manipulated, the Bayesian log evidence which they approximate as the log determinant of the Hessian $\prod_{i}\log(\lambda_{i}/\gamma)$ is invariant to such reparameterisation, where $\gamma$ is the $L2$ regularisation co-efficient. Typically in training the $\gamma$ co-efficient is fixed and hence the log determinant of the Hessian (if it exists) and evidence is subject to the same manipulations \cite{mackay1992bayesian}.
\citet{neyshabur2017exploring} show that it captures generalization for large but not small networks. \citet{jiang2019fantastic} conduct extensive empirical analysis and show that the spectral norm is strongly negatively correlated with generalization and its size is strongly correlated with network depth. However typical comparisons of sharpness are done on the same network, so this study does not conclusively show that for the same network, having a larger spectral norm correlates with improved generalistion. Furthermore, they do not consider the trace or Frobenius norm of the Hessian. \citet{he2019asymmetric} argue that asymmetry in the loss landscape can give the illusion of flat and sharp minima. 
Given the paramount importance of discovering out of sample robust deep learning solutions and the extent to which the concept of flatness has played a pivotal role in the communities discussion, further investigation and theoretical understanding is required to the relevance of sharpness and generalization.

\section{Motivation: beyond reparameterisation \& the deep linear exponential loss}
\label{sec:motivation}
\citet{dinh2017sharp} argue that Hessian based measures can be manipulated from sharp to flat and vice versa without altering the loss. What is rarely discussed is that this also holds true for the gradient. To show this explicitly and simply let us consider the exponential loss $L$, of a deep linear network of $3$ parameters $[w_{1},w_{2},w_{3}]$ for a single data-point $X$ (we generalise this argument to feed forward networks of arbitrary depth with the cross entropy loss in Section \ref{sec:theory}). This model and extensions thereof will serve as our initial intuition for more complicated models later in this paper.

$L = \exp{(w_{1}w_{2}w_{3})X}$. Now the gradient and its norm are given by $\nabla L, ||\nabla L||$ respectively
\begin{equation}
\label{eq:gradnorm}
\begin{aligned}
& \nabla L = (w_{2}w_{3}\delta \hat{w_{1}}+w_{1}w_{3}\delta \hat{w_{2}}+w_{2}w_{1}\delta \hat{w_{3}})X\exp{(w_{1}w_{2}w_{3})X} \\
& ||\nabla L|| = |X|\sqrt{w_{2}^{2}w_{3}^{2}+w_{1}^{2}w_{3}^{2}+w_{2}^{2}w_{1}^{2}}\exp{(w_{1}w_{2}w_{3})X} \\
\end{aligned}
\end{equation}
and the Hessian, its trace and extremal eigenvalue $\mH,\Tr(\mH), \lambda_{max}(\mH)$
\begin{equation}
\mM = \begin{bmatrix}
w_{2}^{2}w_{3}^{2} & w_{1}w_{2}w_{3}^{2}   & w_{2}^{2}w_{1}w_{3}\\
w_{1}w_{2}w_{3}^{2} & w_{1}^{2}w_{3}^{2}   & w_{1}^{2}w_{2}w_{3}\\
w_{2}^{2}w_{1}w_{3} & w_{1}^{2}w_{2}w_{3} & w_{1}^{2}w_{2}^{2}
\end{bmatrix}|X|\exp{(w_{1}w_{2}w_{3})X} 
\end{equation}
\begin{equation}
\label{eq:hesstraceandmax}
\begin{aligned}
& \Tr(\mH) = \lambda_{max}(\mH) = (w_{2}^{2}w_{3}^{2}+w_{2}^{2}w_{1}^{2}+w_{1}^{2}w_{3}^{2})|X|\exp{(w_{1}w_{2}w_{3})X} \\
% & \lambda_{max}(\mH) = (\frac{1}{w_{1}}+\frac{1}{w_{2}}+\frac{1}{w_{3}})\exp{(w_{1}w_{2}w_{3})X} \\
\end{aligned}
\end{equation}
\paragraph{The loss is invariant to rescaling, the gradient and Hessian are not.} For all pairs of transformations $w_{i} \rightarrow \alpha w_{i} \thinspace \& \thinspace w_{j} \rightarrow w_{j}/\alpha$, from the definition of $L$ the loss is unchanged. Neither the Hessian or its Trace are invariant under this transformation. Hence flat minima can be mapped to arbitarily sharp minima without altering the functional output and hence generalisation properties. However, as can be noted from \eqref{eq:gradnorm} and Theorem 3 in \citep{dinh2017sharp} any gradient with norm $0 < ||\nabla L|| \leq \epsilon$, can under the same functional re-parameterisation be transformed into a gradient of arbitrary magnitude and direction. For example consider letting $w_{2} \rightarrow \alpha w_{2}$ and $w_{3} \rightarrow 1/\alpha w_{3}$, then as $\alpha \rightarrow \infty$ the gradient becomes completely aligned in the direction $\delta \hat{w_{3}}$ and tends to infinite magnitude. Hence arguments against Hessian based measures of sharpness due to their instability wrt to reparameterisation, can be equally applied against the use of gradient based optimisation methods. In practice, gradient methods find good solutions and we do not vary the parameterisation at any point in training. Hence the question as to whether Hessian based sharpness measures are good predictors of generalisation, as is commonly believed in the literature is still an open question and requires further investigation.

\paragraph{The Low rank nature of the Hessian:}
Previous empirical works \citep{ghorbani2019investigation,papyan2018full,sagun2016eigenvalues,sagun2017empirical,chaudhari2016entropysgd} have noted a large portion of zero or near zero eigenvalues in the Hessian of deep neural networks. As is noted from equation \eqref{eq:hesstraceandmax}, the largest eigenvalue is equal to the trace and so the matrix is of rank $1$. It can be seen that increasing the product chain does not increase the rank. We extend this line of reasoning in section \ref{sec:theory} to feed forward neural networks using the cross entropy loss to show that the rank is bounded by the number of classes multiplied by the number of neurons times a constant. For large networks, this is usually far smaller than the total number of parameters.  
\paragraph{Smaller losses imply flatter Hessia:} Another interesting aspect of \eqref{eq:hesstraceandmax} is that the trace and maximum eigenvalue are a polynomial function of the weights and the loss is an exponential in the weights. As the loss is driven towards $0$ we expect the exponential to dominate the polynomial. This implies that methods to reduce the weight magnitude, such as $L2$ regularisation, should increase Hessian based measures of sharpness. We show that this is the case experimentally.

\section{Theory}
\label{sec:theory}
In this section we extend our intuition developed under the deep linear network with exponential loss, to feed forward neural networks under the cross entropy loss. The key results of this section, are that we expect Hessian based sharpness metrics to decrease in magnitude as the training loss decreases. Furthermore, we expect the Hessian of a feed forward neural network to be low rank.
\subsection{Similarity between Exponential and Cross Entropy Loss}
\label{subsec:exploss}
% \begin{figure}[h!]
%     \centering
%     \includegraphics[height=2.5cm]{ffnet.png}
%     \caption{Feed Forward NN}
%     \label{fig:ffnn}
% \end{figure}

For the commonly employed cross entropy loss $\ell(h(\vx;\vw), \vy)$.
\begin{equation}
\label{eq:crossentropy}
    \ell(h(\vx_{i};\vw),\vy_{i}) = - \sum_{c}^{d_{\vy}}(1-\mathbbm{1}[\hbar(\vx_{i};\vw)_{c} \neq y_{c}])\log h(\vx_{i};\vw)_{c}
\end{equation}
typically coupled with softmax activation at the final layer, with which we exclusively concern ourselves in this paper. 
\begin{equation}
\label{eq:softmax}
h(\vx_{i},\vw)_{c} = \sigma(\vz)_{c} =  \frac{\exp^{\vz(\vx_{i};\vw)_{c}}}{\sum_{k=1}^{d_{y}}\exp^{\vz(\vx_{i};\vw)_{k}}}
\end{equation} 
Where $d_{y}$ is the number of classes and $\mathbbm{1}$ is the indicator function which takes the value of $1$ for the incorrect class and $0$ for the correct class, $\vz(\vx_{i};\vw)$ is the softmax input. Denoting $L$ as the empirical loss, $q(i)$ as the correct class for sample $i$ 
% and taking the motivational example of the $3$ parameter deep linear neural network, we have 
\begin{equation}
\label{eq:crossentropy}
\begin{aligned}
 & L  = \frac{1}{N}\sum_{i}^{N}\ell(h(\vx_{i};\vw),\vy_{i}) = -\frac{1}{N}\sum_{i}^{N}\sum_{c=1}^{d_{y}}(1-\mathbbm{1}[\hbar(\vx_{i};\vw)_{c} \neq y_{c}])\log h(\vx_{i};\vw)_{c} \\
 & = \frac{1}{N}\sum_{i}^{N}\log\bigg(1+\sum_{k \neq q(i)}^{d_{y}}\exp(h_{k}-h_{q(i)})\bigg) \approx \sum_{i}^{N}\frac{\sum_{k \neq q(i)}^{d_{y}}\exp{[h(\vx_{i},\vw)_{k}-h(\vx_{i},\vw)_{q(i)}]}}{N} \\
% & L =  -\frac{1}{N}\sum_{i}^{N}\sum_{c\neq q(i)}^{d_{y}}(w_{1}w_{2}w_{3}x_{i})_{c}-\log(\sum_{k=1}^{d_{y}}\exp{(w_{1}w_{2}w_{3}x_{i})_{k}}) \\
% & L =  \frac{1}{N}\sum_{i}^{N}\sum_{c\neq q(i)}^{d_{y}}\log(1+\frac{\sum_{k \neq c}^{d_{y}}\exp{(w_{1}w_{2}w_{3}x_{i})_{k}}}{\exp{(w_{1}w_{2}w_{3}x_{i})_{c}}}) \approx \frac{1}{N}\sum_{i}^{N}\sum_{c\neq q(i)}^{d_{y}}\frac{\sum_{k \neq c}^{d_{y}}\exp{(w_{1}w_{2}w_{3}x_{i})_{k}}}{\exp{(w_{1}w_{2}w_{3}x_{i})_{c}}} \\ 
\end{aligned}
\end{equation}
Where for low loss values, we simply taylor expand, obtaining the desired exponential form. In this regime, the motivational arguments of section \ref{sec:motivation} hold. Without approximation, we can use the chain rule and absorbing the numerator of \eqref{eq:softmax} in the denominator, and further rewriting $h(\vx_{i},\vw)_{i}-h(\vx_{i},\vw)_{c} = h_{i,c}$ to unclutter the notation, we have 
\begin{equation}
\label{eq:crossentlossexpanded}
    \begin{aligned}
    & \frac{\partial \ell(h(\vx_{i};\vw),\vy_{i})}{\partial w_{j}} = \frac{-\sum_{i \neq c}\frac{\partial h_{i,c}}{\partial w_{j}}\exp(h_{i,c})}{1+\sum_{i\neq c}\exp(h_{i,c})}
     \\
    & \frac{\partial^{2} \ell(h(\vx_{i};\vw),\vy_{i})}{\partial w_{j}\partial w_{k}} = \frac{-\sum_{i\neq c}\exp(h_{i,c})[\frac{\partial^{2}h_{i,c}}{\partial w_{j}\partial w_{k}}+\frac{\partial h_{i,c}}{\partial w_{j}}\frac{\partial h_{j,c}}{\partial w_{k}}]}{1+\sum_{i\neq c}\exp(h_{i,c})}+\frac{\sum_{i \neq c, l\neq c}\exp(h_{i,c})\frac{\partial h_{i}}{\partial w_{j}}\exp(h_{l,c})\frac{\partial h_{l,c}}{\partial w_{k}}}{(1+\sum_{i\neq c}\exp(h_{i,c}))^{2}}\\
    \end{aligned}
\end{equation}

and hence under the same condition that the loss in \eqref{eq:crossentlossexpanded} tends to $0$, i.e $\exp(h_{i,c})$ goes to $0, \forall i$, both the gradient and the hessian also tend to $0$. In practice, as the output per class cannot go to infinity under finite numerical precision, neither the loss, gradient nor Hessian can go to zero. By assuming differentiability into the softmax input $h(\vx_{i},\vw)$ and expanding as a power series in the weights $\vw$, the Hessian will be given by a polynomial multiplied by an exponential in the weights. As the weights grow in size in order to reduce the loss, because the exponential goes to zero faster than any polynomial grows to infinity, we expect the magnitude of the Hessian spectral norm and trace to reduce. The combination of this exponential polynomial product can be seen explicitly for motivational example in \eqref{eq:hesstraceandmax}. One regularly employed weight reduction technique which is $L2$ regularisation. Hence we investigate its effect on the spectrum of models trained with the cross entropy loss, with and without $L2$ regularisation in section \ref{sec:l2sharpness}.

\subsection{Hessian of feed forward neural network \& the question of rank degeneracy}
\label{subsec:hessianffnn}

\begin{figure}[h!]
    \centering
     \begin{subfigure}{0.22\linewidth}
        \includegraphics[width=1\linewidth,trim={0 0 0 0},clip]{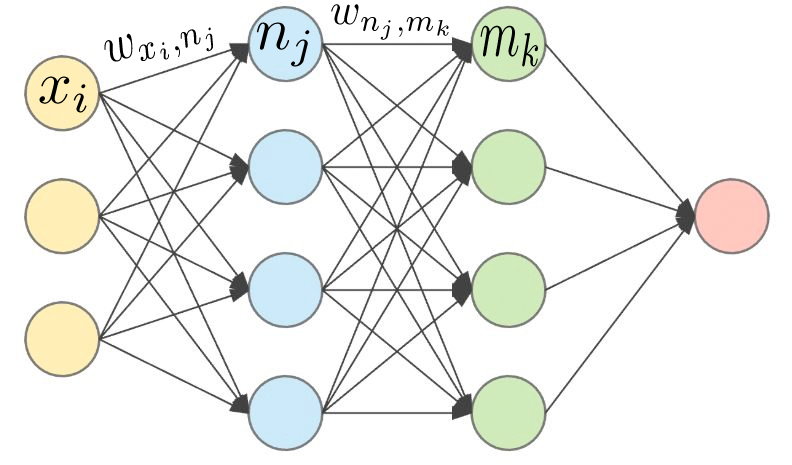}
        \caption{FFNN diagaram}
        \label{subfig:ffnn}
    \end{subfigure}
    \begin{subfigure}{0.24\linewidth}
        \includegraphics[width=1\linewidth,trim={0 0 0 0},clip]{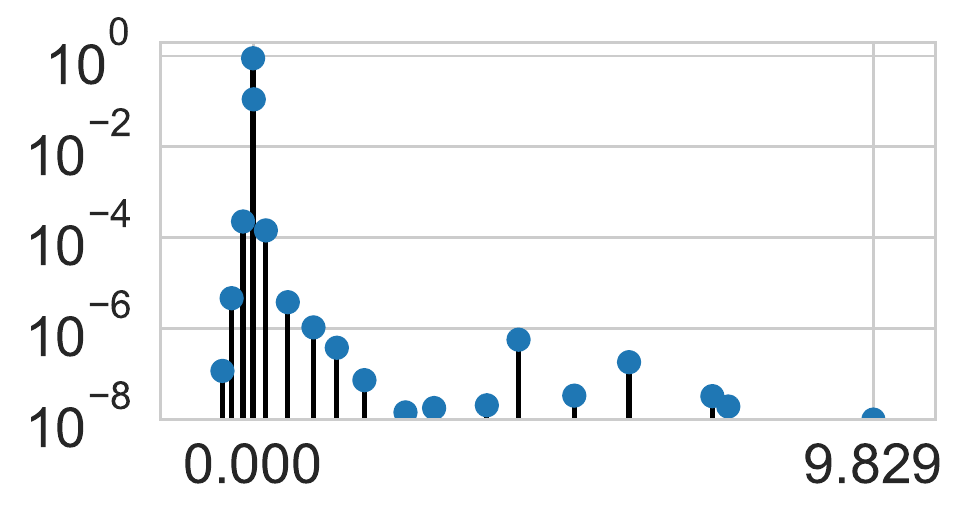}
        \caption{Hessian Epoch $25$}
        \label{subfig:hess25c10vgg16}
    \end{subfigure}
    \begin{subfigure}{0.24\linewidth}
        \includegraphics[width=1\linewidth,trim={0 0 0 0},clip]{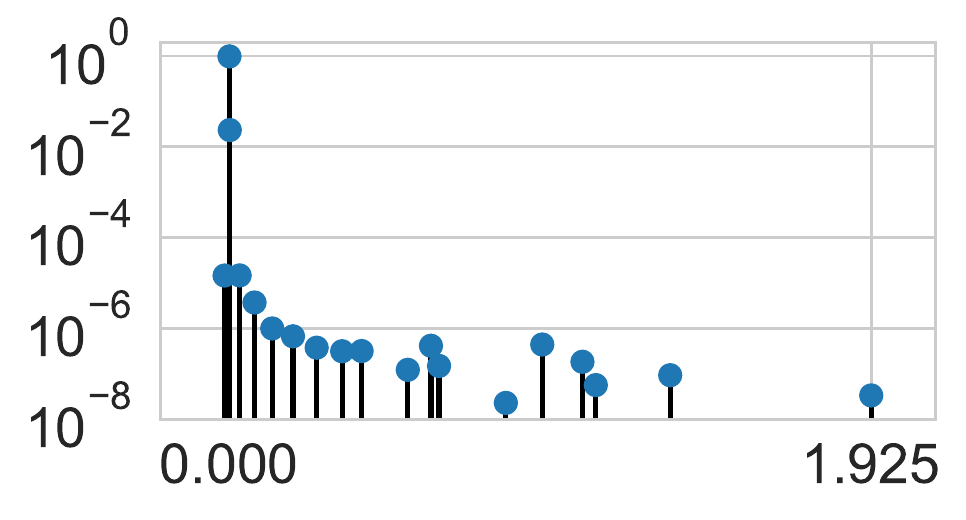}
        \caption{Hessian Epoch $300$}
        \label{subfig:hess300c10vgg16}
    \end{subfigure}
     \begin{subfigure}{0.24\linewidth}
        \includegraphics[width=1\linewidth,trim={0 0 0 0},clip]{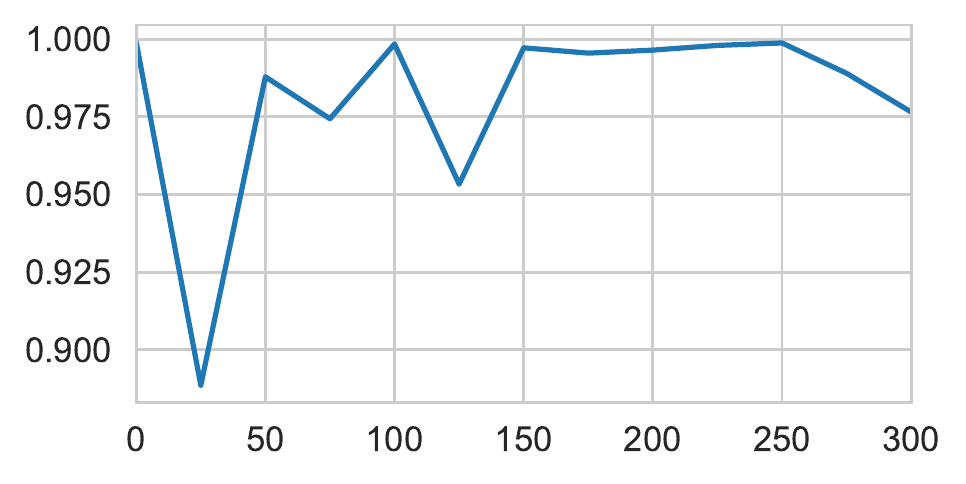}
        \caption{Rank Degeneracy}
        \label{subfig:rankdegen}
    \end{subfigure}
    \caption{Feed forward neural net illustration and Hessian spectrum for VGG-$16$ at various training epochs on the CIFAR-$10$ dataset, along with rank degeneracy at the origin throughout training}\label{fig:vgg16c10}
\end{figure}

% \begin{figure}[h!]
%     \centering
%     \includegraphics[height=3cm]{ffnet.png}
%     \caption{feed forward NN}
%     \label{fig:ffnn}
% \end{figure}
For a feed forward neural network, by considering the paths as shown in Figure \ref{subfig:ffnn}, it can be seen that the input to neuron $m_{k}$ is given by $\sum_{n_{j}}^{N_{1}}\sum_{i}^{d_{x}}x_{i}\vw_{x_{i},n_{j}}\vw_{n_{j},m_{k}}$, where $N_{1}$ is the number of neurons in layer $1$. Generalising this procedure we find for a neural network $d-1$ hidden layers
\begin{equation}
    h_{m} = \prod_{l=1}^{d-1}\sum_{n_{i,l}=1}^{N_{i,l}}\sum_{i}^{d_{x}}\vx_{i}\vw_{n_{i,l},n_{i,l+1}}\delta(n_{i,d}=m)
\end{equation}
where $n_{i,l_{1}} = x_{i}$ and we fix to the desired output class, through the delta function. The Hessian of the loss in the small loss limit tends to
\begin{equation}
        \frac{\partial^{2} \ell(h(\vx_{i};\vw),\vy_{i})}{\partial w_{\phi,\kappa}\partial w_{\theta,\nu}} \rightarrow -\sum_{m\neq c}\exp(h_{m})\bigg[\frac{\partial^{2}h_{m}}{\partial w_{\phi,\kappa}\partial w_{\theta,\nu}}+\frac{\partial h_{m}}{\partial w_{\phi,\kappa}}\frac{\partial h_{m}}{\partial w_{\theta,\nu}}\bigg]
\end{equation}
\begin{equation}
\label{eq:expandingloss}
\begin{aligned}
    & \bigg[\frac{\partial^{2}h_{m}}{\partial w_{\phi,\kappa}\partial w_{\theta,\nu}}+\frac{\partial h_{m}}{\partial w_{\phi,\kappa}}\frac{\partial h_{m}}{\partial w_{\theta,\nu}}\bigg] =  \prod_{l=1}^{d-1}\sum_{n_{i,l}\neq [(\phi, \kappa),(\theta, \nu)]}^{N_{i,l}}\sum_{i}^{d_{x}}\vx_{i}\vw_{n_{i,l},n_{i,l+1}}\delta(n_{i,d}=m) \\
    & +\bigg(\prod_{l=1}^{d-1}\sum_{n_{i,l}\neq (\theta, \nu)}^{N_{i,l}}\sum_{i}^{d_{x}}\vx_{i}\vw_{n_{i,l},n_{i,l+1}}\delta(n_{i,d}=m)\bigg)
    \bigg(\prod_{l=1}^{d-1}\sum_{n_{j,l}\neq (\phi, \kappa)}^{N_{j,l}}\sum_{i}^{d_{x}}\vx_{i}\vw_{n_{j,l},n_{j,l+1}}\delta(n_{j,d}=m)\bigg)
    \\
    \end{aligned}
\end{equation}
Each product of weights contributes an object of rank-$1$ (as shown in section \ref{sec:motivation}). Furthermore, the rank of a product is the minimum of the constituent ranks, i.e $\text{rank}(AB) = \min \text{rank}(A,B)$. Hence \eqref{eq:expandingloss} is rank bounded by a  $2(\sum_{l}N_{l} + d_{x})$, where  $N_{l}$ is the total numbers of neurons in the network. By rewriting the loss per-sample and repeating the same arguments and including the class factor
\begin{equation}
    \frac{\partial^{2}\ell}{\partial w_{k}\partial w_{l}} = - \frac{\partial^{2}h_{q(i)}}{\partial w_{k}\partial w_{l}} + \frac{\sum_{j}\exp(h_{j})\sum_{i}\exp(h_{i})(\frac{\partial^{2}h_{i}}{\partial w_{k} \partial w_{l}}+\frac{\partial h_{i}}{\partial w_{k}}\frac{\partial h_{i}}{\partial w_{l}})-\sum_{i}\exp(h_{i})\frac{\partial h_{i}}{\partial w_{k}}\sum_{j}\frac{\partial h_{j}}{\partial w_{l}}\exp(h_{j})}{[\sum_{j}\exp(h_{j})]^{2}}
\end{equation}
We obtain a rank bound of $4d_{y}(\sum_{l}N_{l} + d_{x})$. To give some context, along with a practical application of a real network and dataset, for the CIFAR-$10$ dataset, the VGG-$16$ \cite{simonyan2014very} contains $1.6 \times 10^{7}$ parameters, the number of classes is $10$ and the total number of neurons is $13,416$ and hence the bound gives us a spectral peak at the origin of at least $1-\frac{577,600}{1.6\times 10^{7}} = 0.9639$. In order to validate this in practice, as we cannot eigendecompose any real neural network Hessia, we use the Lanczos algorithm implementation to get a moment matched spectral approximation \citep{granziol2019mlrg} and take the smallest Ritz value to be the origin \footnote{We note that taking more or less Lanczos steps can alter the result and if there is an eigenvalue near the origin, the weight may split giving a lower value}. As shown in Figure \ref{subfig:rankdegen}, this bound is largely respected in practice and where it is broken, this is because there is a Ritz value very close to the smallest at the origin, as shown in Figure \ref{subfig:hess25c10vgg16}. When there is a sufficient gap between the Ritz values, such as in Figure \ref{subfig:hess300c10vgg16} the rank degeneracy exceeds the bound.

\section{Weight Decay and Sharpness}
\label{sec:l2sharpness}

$L2$ regularisation is a well known trick of the trade, it is regularly used to help generalisation, having been showed to reduce the effect of static noise on the target \citep{krogh1992simple}, furthermore low weight norm solutions have been argued to help generalisation \cite{wilson2017marginal}. In this section we test the intuition from section \ref{sec:theory} that $L2$ regularisation should increase the sharpness of Hessian based measures. We find this to be the case for Logistic Regression, multi layer perceptrons, small convolutional neural networks and Wide Residual networks. For further experimental details, such as the learning rate schedule (linear decay) employed and the finer details of the spectral visualisation method see Appendix \ref{sec:experimentdetails}. Where the Hessian is positive definite we report the trace and where not we report the Frobenius norm as an alternative measure to the spectral norm.

\begin{figure}[h!]
    \centering
     \begin{subfigure}{0.32\linewidth}
        \includegraphics[width=1\linewidth,trim={0 0 0 0},clip]{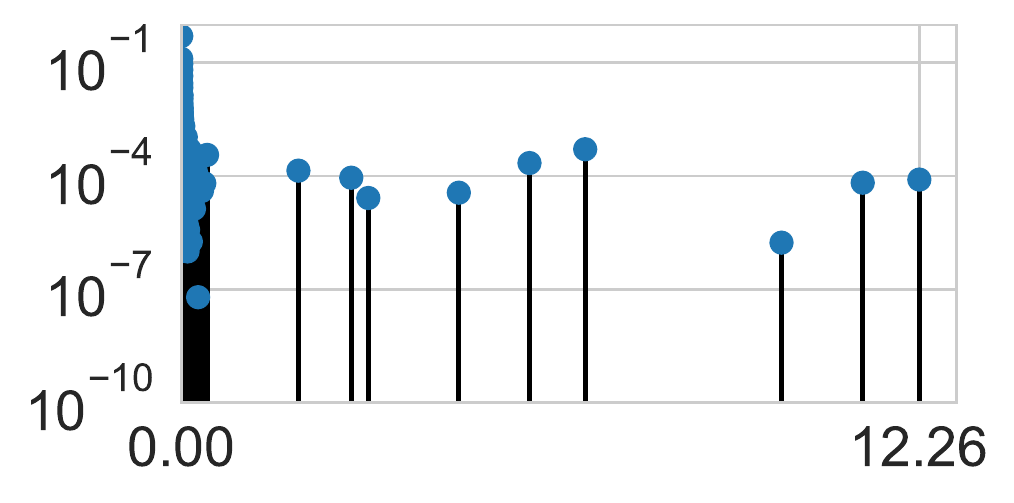}
        \caption{$Acc = 93.48$ $\lambda = 0$}
        \label{subfig:logisticwd0}
    \end{subfigure}
    \begin{subfigure}{0.32\linewidth}
        \includegraphics[width=1\linewidth,trim={0 0 0 0},clip]{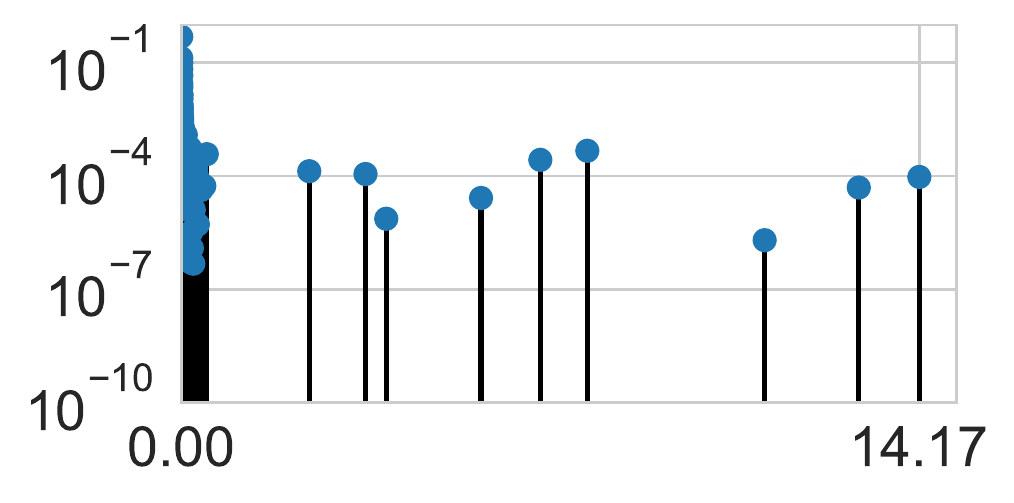}
        \caption{$Acc = 94$  $\lambda = 10^{-4}$}
        \label{subfig:logisticwd0001}
    \end{subfigure}
     \begin{subfigure}{0.32\linewidth}
        \includegraphics[width=1\linewidth,trim={0 0 0 0},clip]{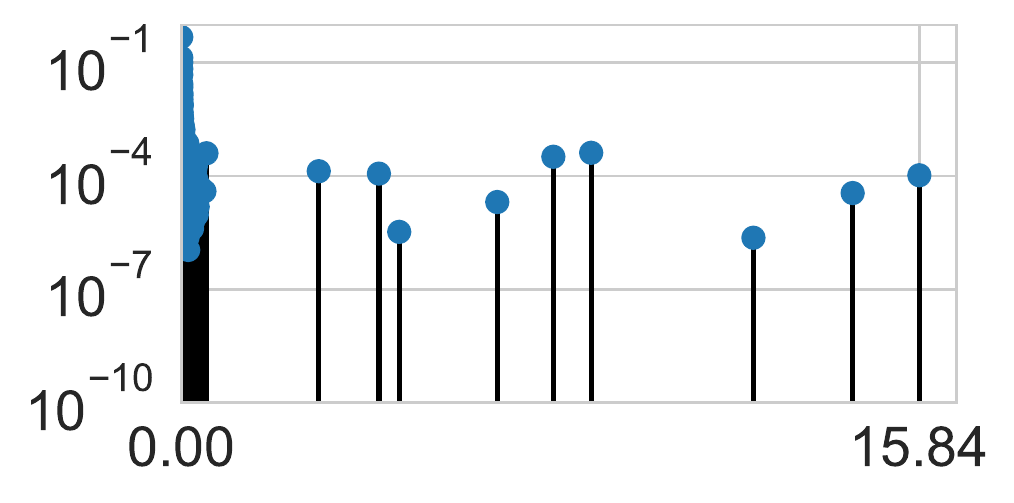}
        \caption{$Acc = 94.08$ $\lambda = 5\times 10^{-4}$}
        \label{subfig:logistic0005}
    \end{subfigure}
    \caption{Hessian spectrum for Logistic regression after $1000$ epochs of SGD on the MNIST dataset, for various $L2$ regularisation co-efficients $\lambda$}
    \label{fig:logisticspec}
    
\end{figure}

\paragraph{Logistic Regression on MNIST:}
The simplest Neural Network model, corresponding to a $0$ hidden layer feed forward neural network, is the multi-class equivalent of Logistic Regression, the Softmax. By the diagonal dominance theorem \citep{cover2012elements} the Hessian of logistic regression is positive semi-definite, so the loss surface is convex, strictly convex with $L2$ regularisation. For a convex objective any local minimum is by definition global and hence there is no complexity in distinguishing between minima. Despite the convexity, there is no analytical solution to the Softmax and hence the system is solved iteratively, using (stochastic) gradient descent. We run Logistic regression on the MNIST dataset \citep{lecun1998mnist}, splitting the training set into $45,000$ training and $5,000$ validation samples. The total parameter count is $7850$. We run for $1000$\footnote{we specifically use an abnormally large number of epochs to make sure that convergence is not an issue} epochs with learning rate and momentum $[0.03,0.9]$ and various levels of $L2$ regularisation on the grid $\lambda \in [0,0.0001,0.0005]$. The validation accuracy increases incrementally with increased weight decay $[93.48,94,94.08]$. We plot the spectra of the final solution in Figure \ref{fig:logisticspec}. We note here that for increasing weight decay co-efficient, which corresponds to higher performing testing solutions, the spectral norm increases, from $\lambda_{1} = [12.26,14.17,15.84]$, the mean eigenvalue also increases from $\mu = [0.0132,0.0153,0.0171]$. This shows that greater Hessian based measures of sharpness, occur for solutions with improved generalisation.

\begin{figure}[h!]
    \centering
     \begin{subfigure}{0.28\linewidth}
        \includegraphics[width=1\linewidth,trim={0 0 0 0},clip]{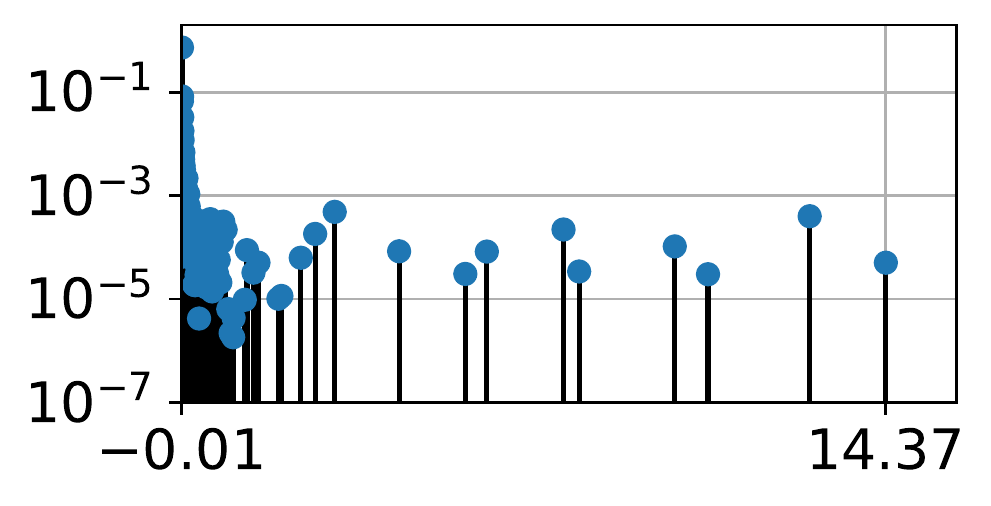}
        \caption{$Acc = 94.4$, $\lambda = 0$}
        \label{subfig:mlpwd0}
    \end{subfigure}
    \begin{subfigure}{0.28\linewidth}
        \includegraphics[width=1\linewidth,trim={0 0 0 0},clip]{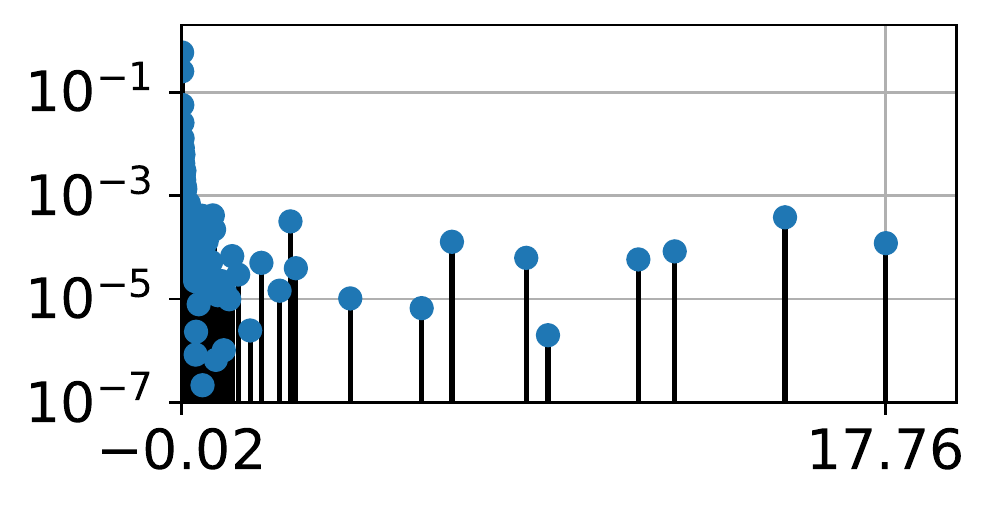}
        \caption{$Acc = 96.7$, $\lambda = 5e^{-4}$}
        \label{subfig:mlpwd0005}
    \end{subfigure}
     \begin{subfigure}{0.37\linewidth}
        \includegraphics[width=1\linewidth,trim={0 0 0 0},clip]{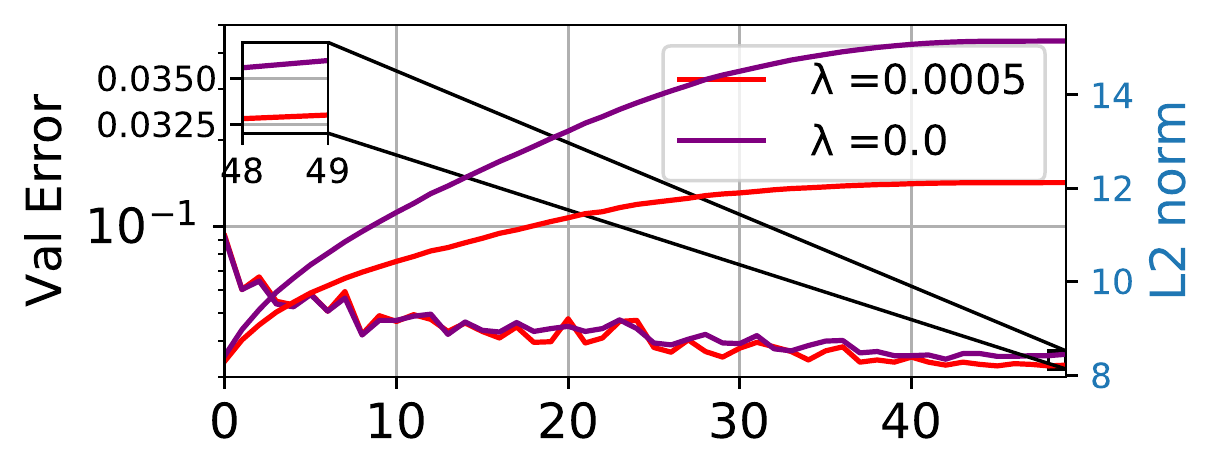}
        \caption{Test Error \& $L2$ norm}
        \label{subfig:mlptest}
    \end{subfigure}
    \caption{Hessian spectrum for MLP after $50$ epochs of SGD on the MNIST dataset, for various $L2$ regularisation co-efficients $\lambda$}
    \label{fig:MLP}
    
\end{figure}

\begin{figure}[h!]
    \centering
     \begin{subfigure}{0.28\linewidth}
        \includegraphics[width=1\linewidth,trim={0 0 0 0},clip]{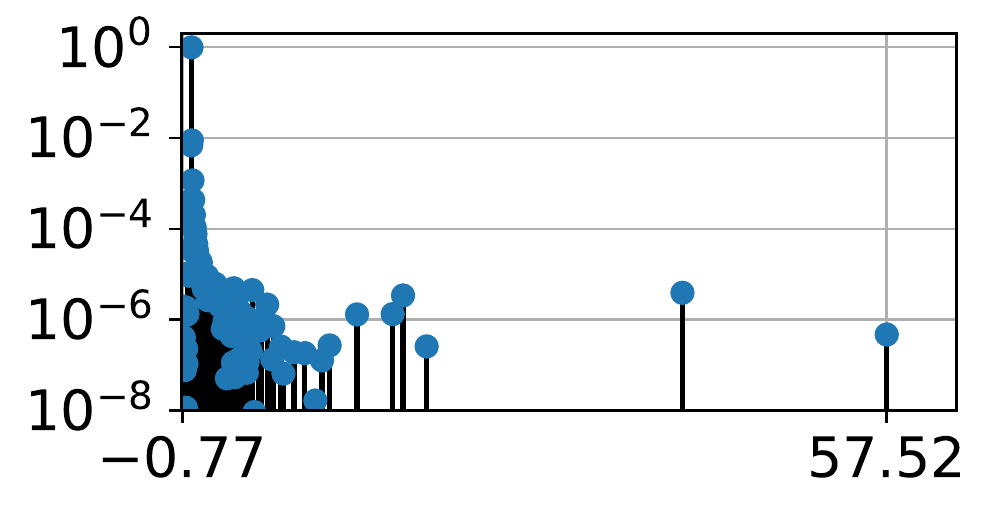}
        \caption{$Val = 57.7$, $\lambda = 0$}
        \label{subfig:cnnwd0}
    \end{subfigure}
    \begin{subfigure}{0.28\linewidth}
        \includegraphics[width=1\linewidth,trim={0 0 0 0},clip]{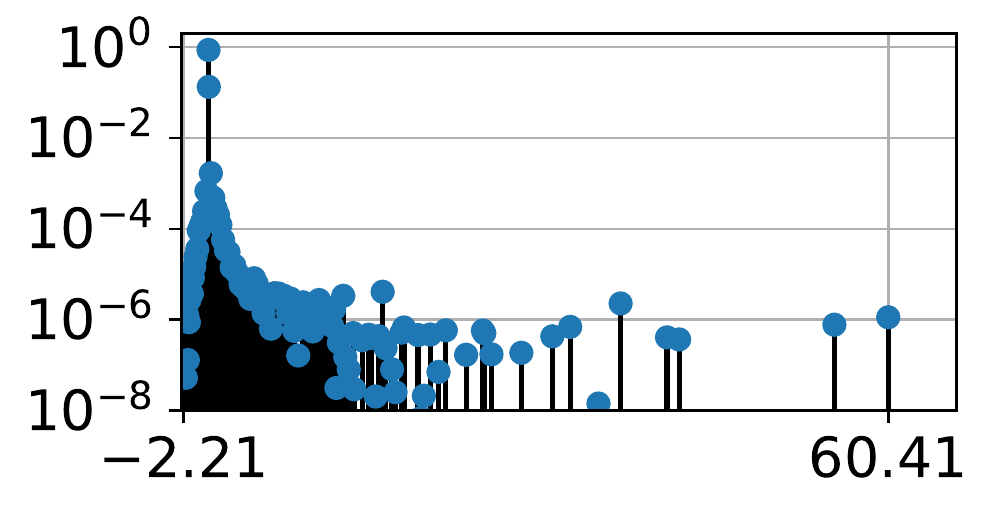}
        \caption{$Val = 62.1$, $\lambda = 5e^{-4}$}
        \label{subfig:cnwd0005}
    \end{subfigure}
     \begin{subfigure}{0.37\linewidth}
        \includegraphics[width=1\linewidth,trim={0 0 0 0},clip]{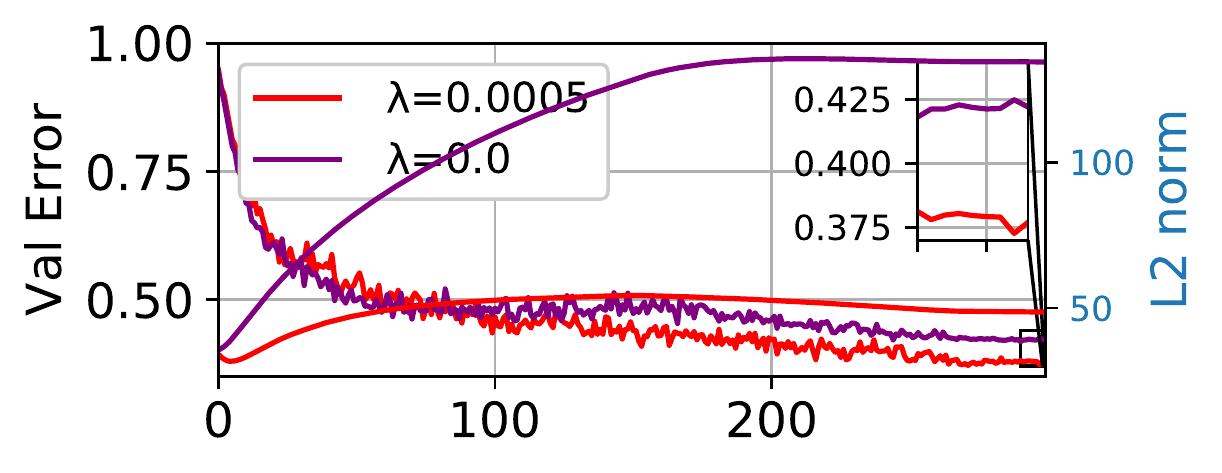}
        \caption{Test Error \& $L2$ norm}
        \label{subfig:cnntest5}
    \end{subfigure}
    \caption{Hessian spectrum for CNN after $300$ epochs of SGD on the CIFAR-$100$ dataset, for various $L2$ regularisation co-efficients $\lambda$}
    \label{fig:cnn}
    
\end{figure}

\begin{figure}[h!]
    \centering
     \begin{subfigure}{0.28\linewidth}
        \includegraphics[width=1\linewidth,trim={0 0 0 0},clip]{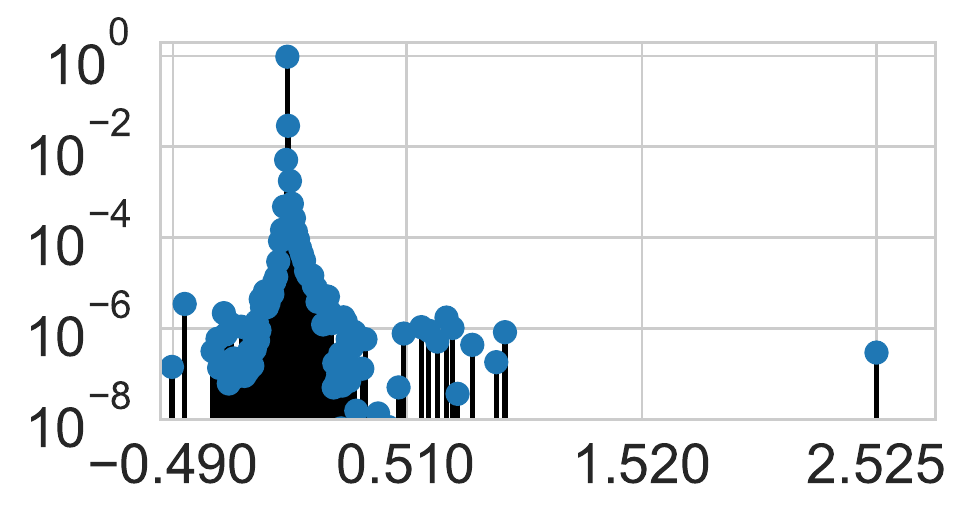}
        \caption{$Val = 73.12$, $\lambda = 0$}
        \label{subfig:p164wd0}
    \end{subfigure}
    \begin{subfigure}{0.28\linewidth}
        \includegraphics[width=1\linewidth,trim={0 0 0 0},clip]{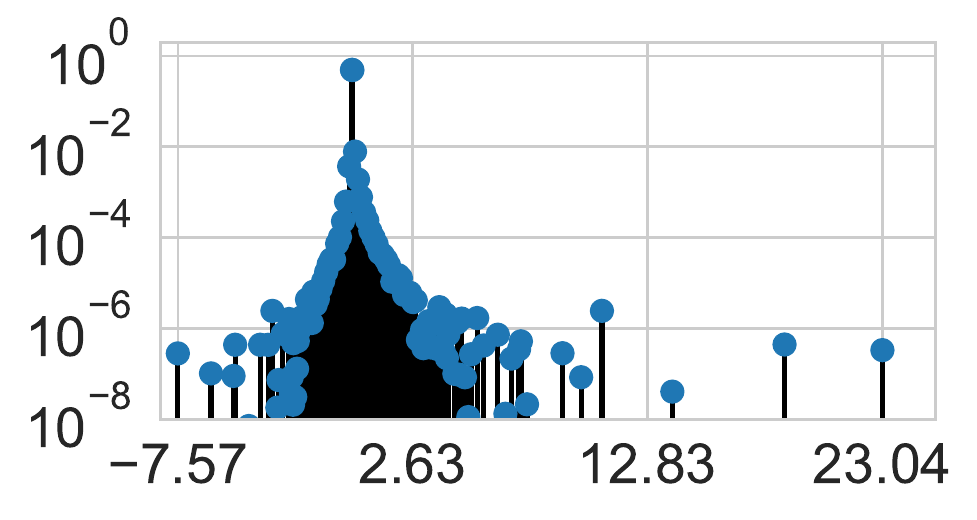}
        \caption{$Val = 77.36$, $\lambda = 5e^{-4}$}
        \label{subfig:p164wd0005}
    \end{subfigure}
     \begin{subfigure}{0.37\linewidth}
        \includegraphics[width=1\linewidth,trim={0 0 0 0},clip]{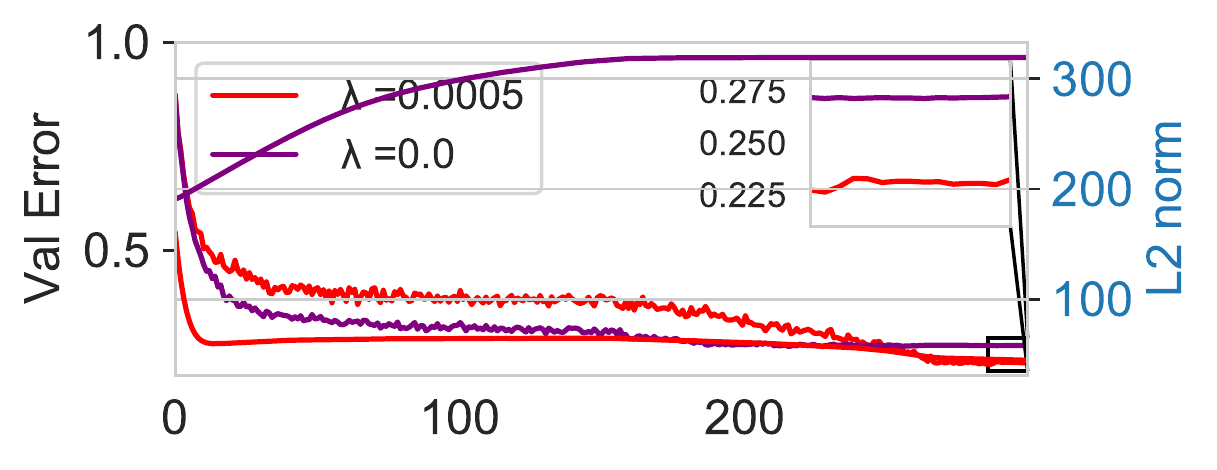}
        \caption{Test Error \& $L2$ norm}
        \label{subfig:p164test5}
    \end{subfigure}
    \caption{Hessian spectrum for PreResNet-$164$ after $300$ epochs of SGD on the CIFAR-$100$ dataset, for various $L2$ regularisation co-efficients $\lambda$}
    \label{fig:p164}
    \vspace{-20pt}
\end{figure}

\begin{figure}[h!]
    \centering
     \begin{subfigure}{0.28\linewidth}
        \includegraphics[width=1\linewidth,trim={0 0 0 0},clip]{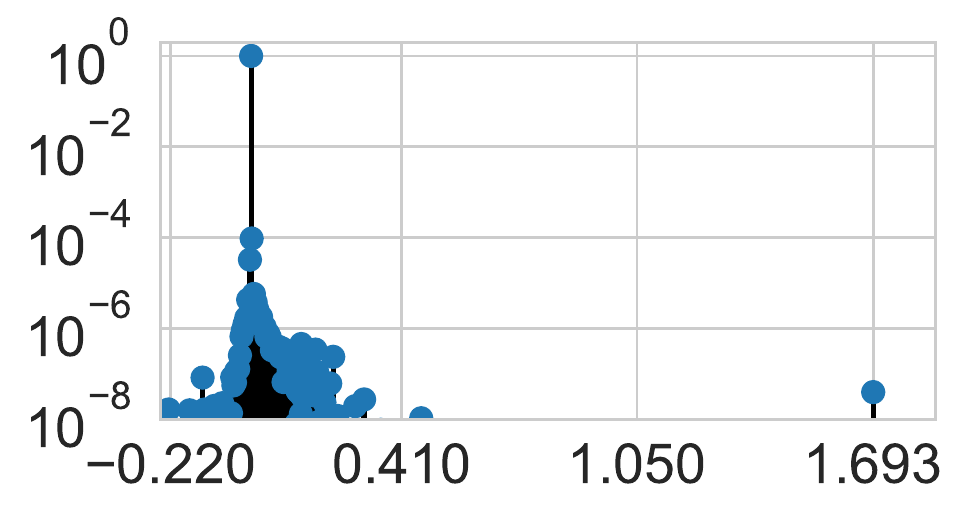}
        \caption{$Val = 75.2$, $\lambda = 0$}
        \label{subfig:wrntwd0}
    \end{subfigure}
    \begin{subfigure}{0.28\linewidth}
        \includegraphics[width=1\linewidth,trim={0 0 0 0},clip]{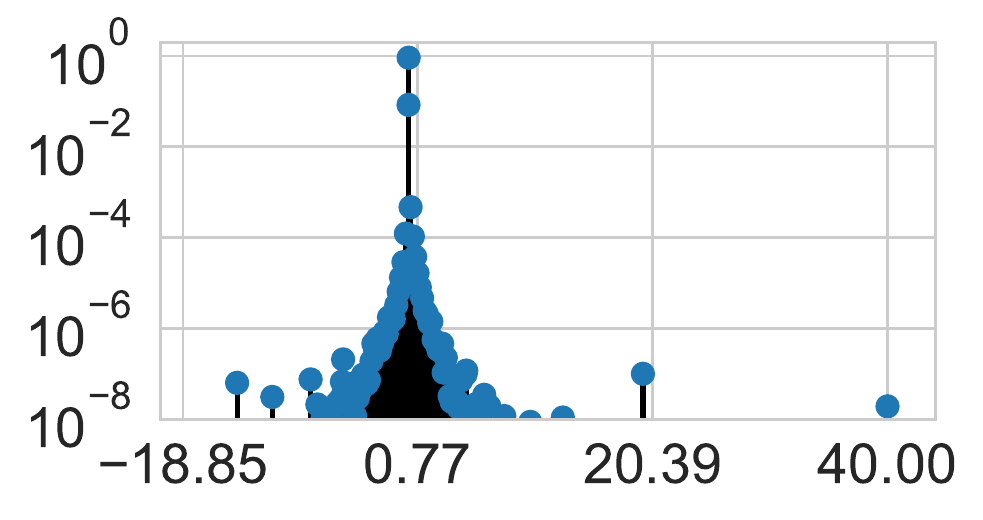}
        \caption{$Val = 80.6$, $\lambda = 5e^{-4}$}
        \label{subfig:wrntwd0005}
    \end{subfigure}
     \begin{subfigure}{0.28\linewidth}
        \includegraphics[width=1\linewidth,trim={0 0 0 0},clip]{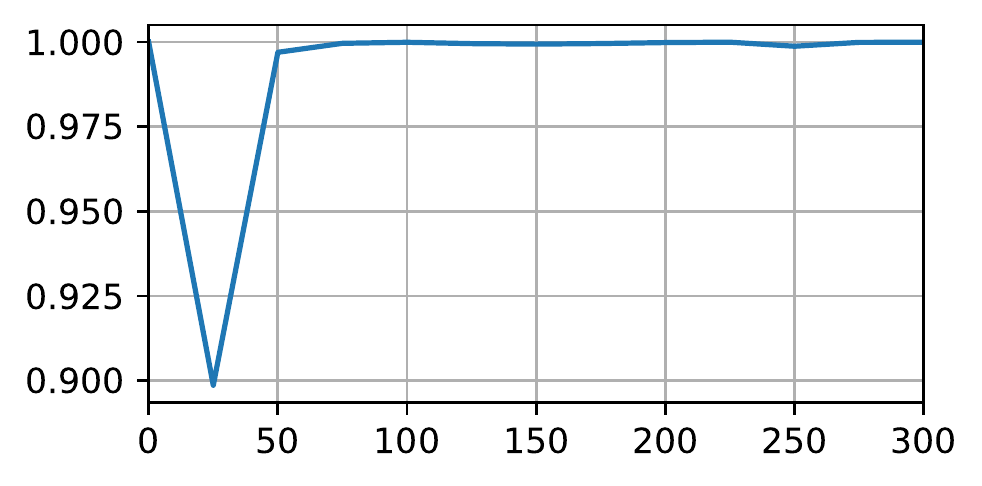}
        \caption{Rank Degeneracy ratio}
        \label{subfig:wrntrankdegenw}
    \end{subfigure}
    \caption{Hessian spectrum for WideResNet$28\times10$ after $300$ epochs of SGD on the CIFAR-$100$ dataset, for various $L2$ regularisation co-efficients $\lambda$, Batch Norm Train mode}
    \label{fig:widresbntrain}
    \vspace{-20pt}
\end{figure}

\begin{figure}[h]
    \centering
     \begin{subfigure}{0.28\linewidth}
        \includegraphics[width=1\linewidth,trim={0 0 0 0},clip]{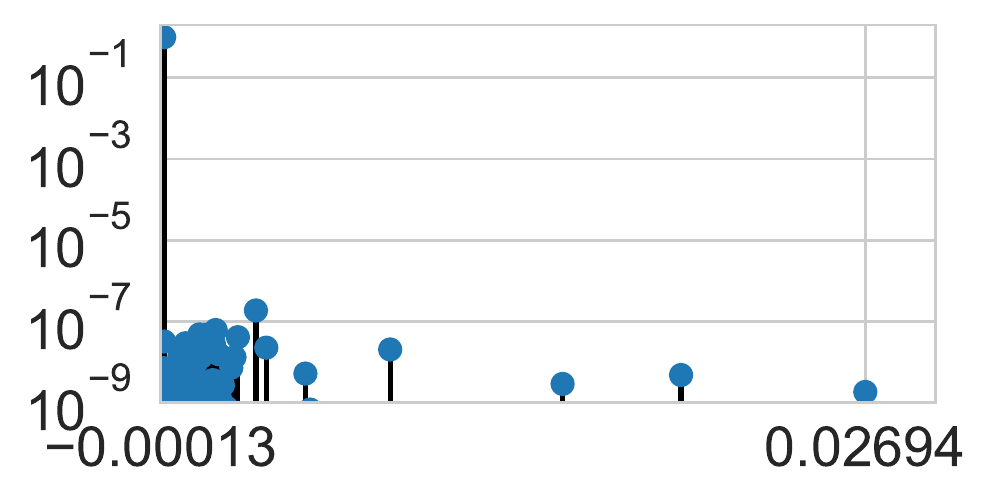}
        \caption{$Val = 75.2$, $\lambda = 0$}
        \label{subfig:wrnwd0}
    \end{subfigure}
    \begin{subfigure}{0.28\linewidth}
        \includegraphics[width=1\linewidth,trim={0 0 0 0},clip]{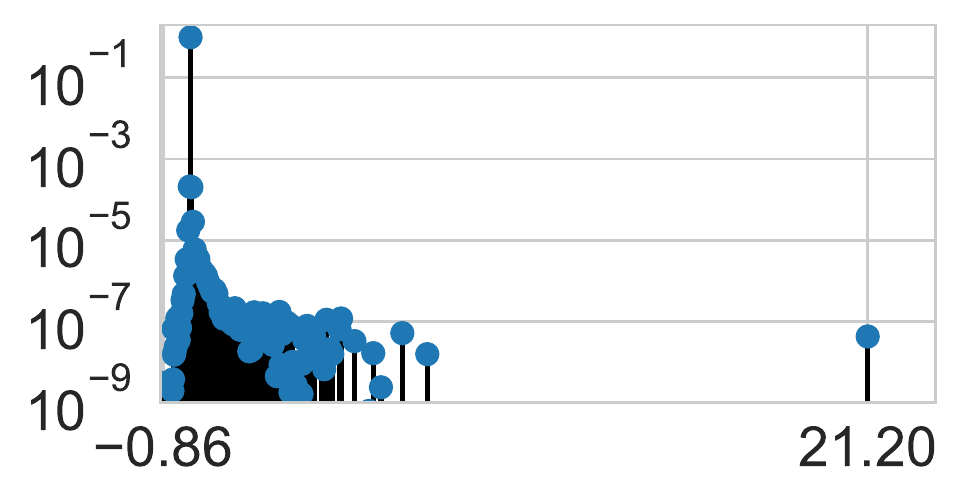}
        \caption{$Val = 80.6$, $\lambda = 5e^{-4}$}
        \label{subfig:wrnwd0005}
    \end{subfigure}
     \begin{subfigure}{0.37\linewidth}
        \includegraphics[width=1\linewidth,trim={0 0 0 0},clip]{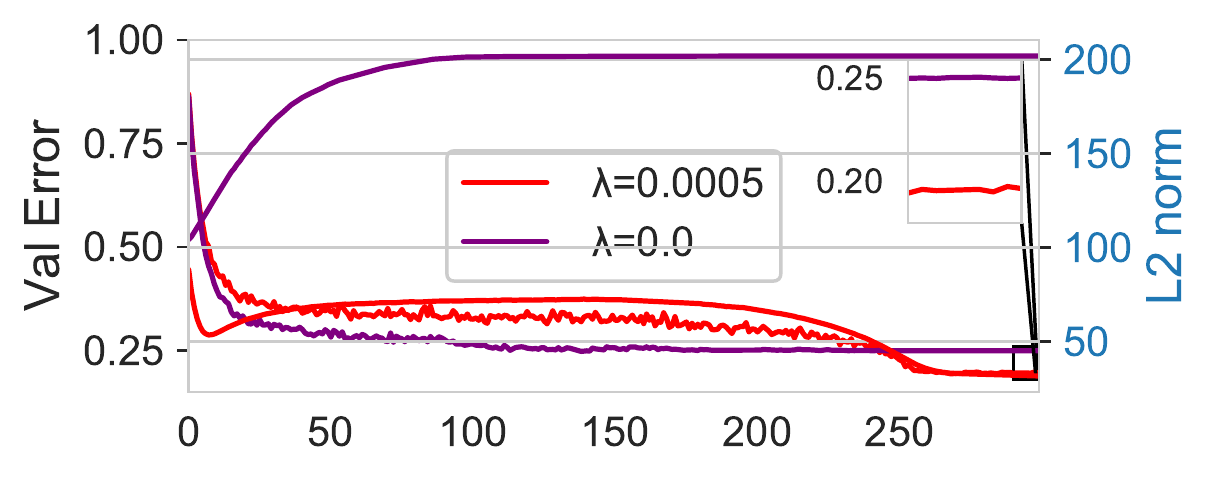}
        \caption{Test Error \& $L2$ norm}
        \label{subfig:wrntest}
    \end{subfigure}
    \caption{Hessian spectrum for WideResNet$28\times10$ after $300$ epochs of SGD on the CIFAR-$100$ dataset, for various $L2$ regularisation co-efficients $\lambda$, batch norm evaluation mode}
    \label{fig:widresbneval}
    
\end{figure}

\paragraph{MLP:}
For the single layer perceptron on the MNIST dataset, with a hidden later of $100$ units, parameter count $9960$, trained for $50$ epochs with an identical schedule. We similarly find as shown in Figure \ref{fig:MLP} that the addition of weight decay both increases the generalisation accuracy (from $94.4$ to $96.7$) but also increases the spectral norm. The Frobenius norm, which we use since the eigenvalues are no longer all positive, hence sharp solutions with negative and positive directions could cancel, is also increased from $0.12$ to $0.16$, indicating a sharper solution. The non regularised solution has a training accuracy of $96.88$ and the regularised solution of $96.65$. We plot both the test error and weight norm in Figure \ref{subfig:mlptest}.
\paragraph{CNN:}
For the $9$ layer convolutional neural network on the CIFAR-$100$ dataset, with parameter count $1,387,108$ trained on the CIFAR-$100$ dataset, with a learning rate of $\alpha = 0.03, \rho = 0.9$ for $300$ epochs. We also observe that adding weight decay increases the spectral norm in Figure \ref{fig:cnn} and the Frobenius norm also from $0.01$ to $0.02$. In this particular case, the differential in training and test loss of the non regularised solution is $1.89$, whereas that of the regularised sharper solution is $1.57$, hence even accounting for different abilities to perform on the training data $[82.2,87.2]$ the sharper solution is still less far away from its training loss estimate that the less sharp solution. We plot the test error and weight norm in Figure \ref{subfig:cnntest5}.
\paragraph{PreResNet-$164$}
For the preactivated residual network on the CIFAR-$100$ dataset with parameter count $1,726,388$, we achieve a training accuracy of $99.89 \%$ with $L2$ regularisation and $99.99 \%$ without. The validation performance is $77.36 \%$ with $L2$ and $73.12 \%$ without. The non regularised solution, is significantly flatter, in spectral norm as shown in Figures \ref{subfig:p164wd0} and \ref{subfig:p164wd0005} and in Frobenius norm, with a value of $5 \times 10^{-5}$ instead of $3 \times 10^{-3}$.
\paragraph{WideResNet-$28\times10$:}
For the wide residual network on the CIFAR-$100$ dataset, with parameter count $36,546,980$, the training accuracy with weight decay is $>99.99 \%$ and $>99.995 \%$ without. However the validation accuracies differ by more than $5 \%$ and consistent with the rest of this paper, the unregularised solution, which performs worse in terms of test accuracy and test loss, is significantly flatter, as shown in Figure \ref{fig:widresbntrain}. The non regularised solution has a spectral norm almost $30\times$ smaller, and the Frobenius norm is $2.3 \times 10^{-7}$ as opposed to the regularised solutions value of $1 \times 10^{-4}$. We plot the rank degeneracy, using the same method as in Section \ref{subsec:hessianffnn}.
\paragraph{How does batch normalisation affect curvature?} %Another point of consideration, which we have not seen explored in the literature, is the variations in curvature calculations using batch normalisation. 
During training both the mean and variance of the batch normalisation layers are adapted to the specific batch, whereas at evaluation they are fixed (to their exponentially moving average). This is done so that the transforms can function even if the prediction set is only $1$ sample\footnote{a $1$ sample set has no variance}. Previous works investigating neural network Hessia \citep{papyan2018full,ghorbani2019investigation} do not consider this free parameter in batch-normalisation and its effect on the spectrum. From a sharpness and generalisation perspective, we would consider that it is the model that is making predictions that we should evaluate. Changing batch normalisation to the evaluation mode, we find that a somewhat different curvature profile, as shown in Figure \ref{fig:widresbneval}. In this case the sharpness of the regularised solution in terms of the spectral norm is nearly $1000$ times larger than that of the regularised, better generalising solution. The Frobenius norm, for the regularised solution is $4.9 \times 10^{-5}$ as opposed to $9.8 \times 10^{-12}$, so $\mathcal{O}(10^{7})$ larger.

\section{Sharpness and Adaptive optimisation}
\begin{figure}[h!]
    \centering
     \begin{subfigure}{0.28\linewidth}
        \includegraphics[width=1\linewidth,trim={0 0 0 0},clip]{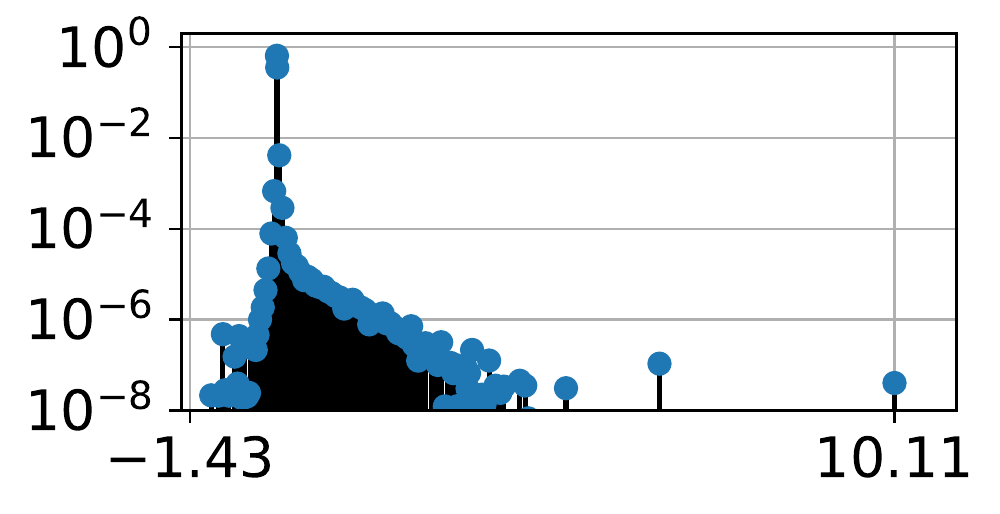}
        \caption{$Val = 73.5$, SGD}
        \label{subfig:vgg16sgdbntrain}
    \end{subfigure}
    \begin{subfigure}{0.28\linewidth}
        \includegraphics[width=1\linewidth,trim={0 0 0 0},clip]{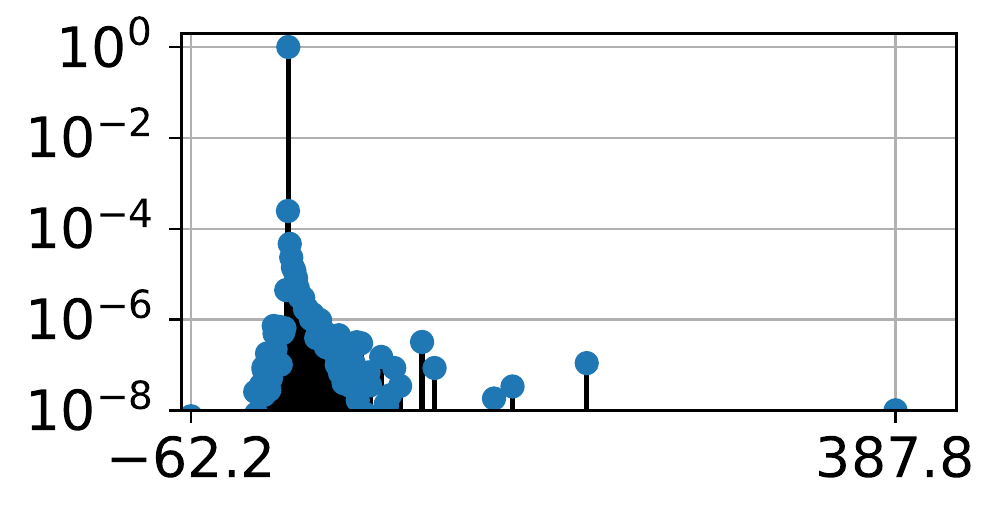}
        \caption{$Val = 75.8$, Gadam}
        \label{subfig:vgg16gadambntrain}
    \end{subfigure}
     \begin{subfigure}{0.37\linewidth}
        \includegraphics[width=1\linewidth,trim={0 0 0 0},clip]{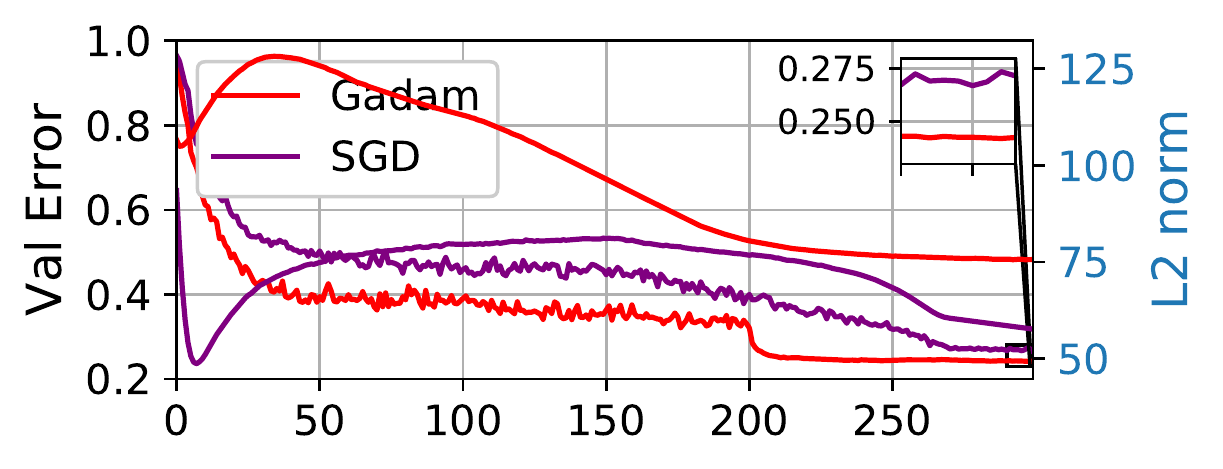}
        \caption{Test Error \& $L2$ norm}
        \label{subfig:vgg16test}
    \end{subfigure}
    \caption{Hessian spectrum for VGG-$16$BN after $300$ epochs of SGD on the CIFAR-$100$ dataset, for various optimisation algorithms [SGD, Gadam], batch norm train mode}\label{fig:gadamspectrain}
\end{figure}

\begin{figure}[h!]
    \centering
     \begin{subfigure}{0.28\linewidth}
        \includegraphics[width=1\linewidth,trim={0 0 0 0},clip]{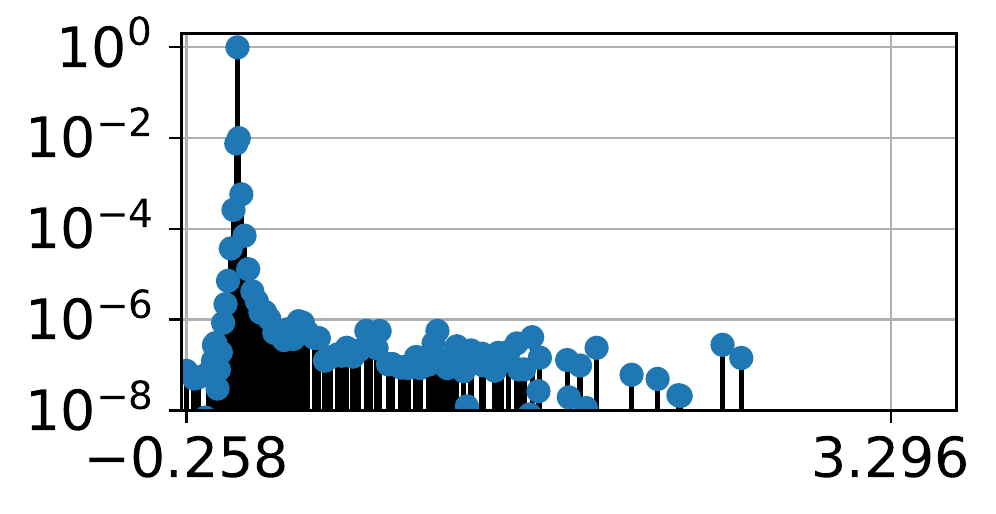}
        \caption{$Val = 94.3$, SGD}
        \label{subfig:sgdtrain}
    \end{subfigure}
    \begin{subfigure}{0.28\linewidth}
        \includegraphics[width=1\linewidth,trim={0 0 0 0},clip]{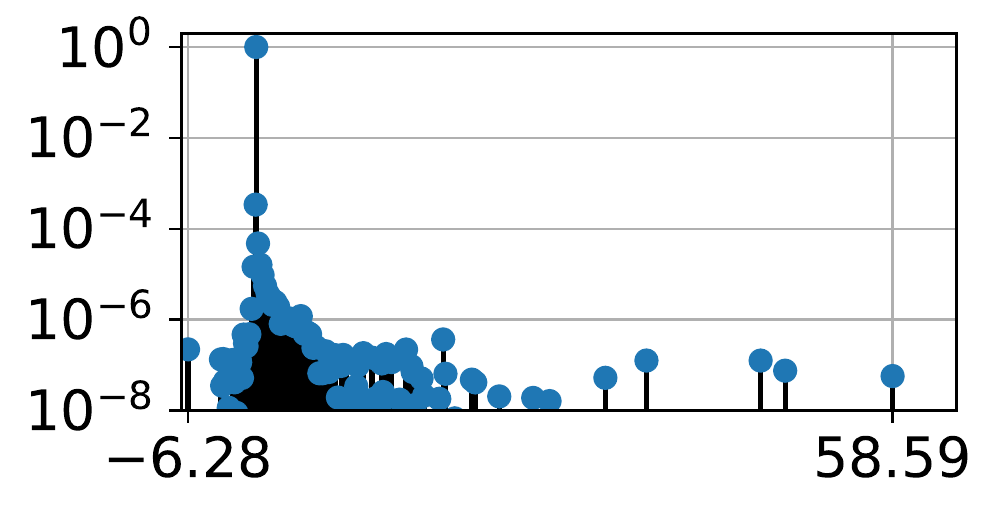}
        \caption{$Val = 95.1$, Gadam}
        \label{subfig:gadamtrain}
    \end{subfigure}
     \begin{subfigure}{0.37\linewidth}
        \includegraphics[width=1\linewidth,trim={0 0 0 0},clip]{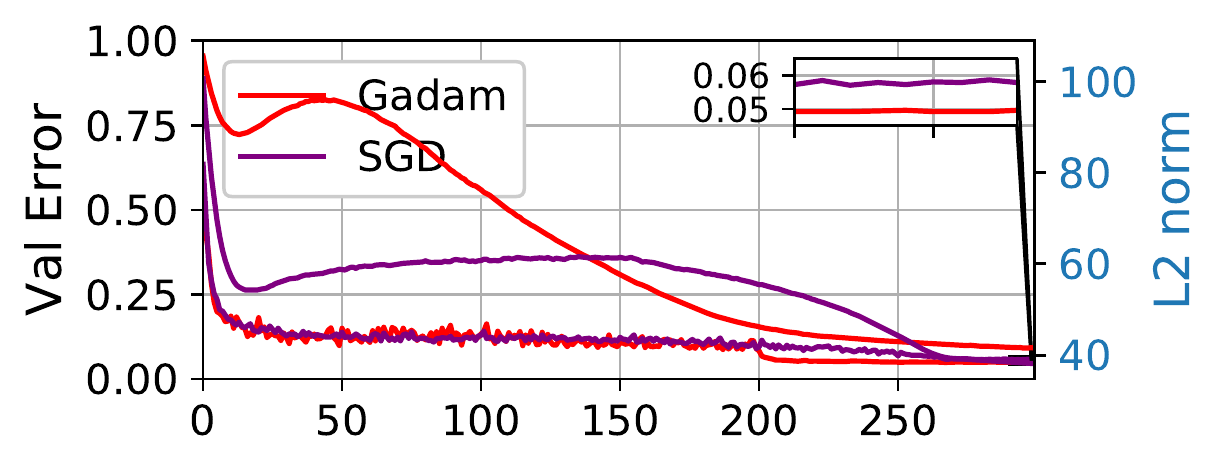}
        \caption{Test Error \& $L2$ norm}
        \label{subfig:testadap}
    \end{subfigure}
    \caption{Hessian spectrum for VGG-$16$BN after $300$ epochs of SGD on the CIFAR-$10$ dataset, for various optimisation algorithms [SGD, Gadam], batch norm train mode}
    \vspace{-20pt}
    \label{fig:gadamc10spectrain}
\end{figure}

Given that all high performing solutions use some form of weight regularisation, we consider whether sharpness can be a useful indicator in the wild for the same neural network trained on the same dataset, but with alternative optimisers and schedules. We use the VGG-$16$ with batch-normalisation on the CIFAR-$10$/$100$ datasets. We use the Gadam optimiser \citep{granziol2020iterate}, which combines Adam \citet{kingma2014adam}, decoupled weight decay and iterate averaging to achieve improved generalisation without compromising adaptivity. We use a decoupled weight decay of $0.35/0.25$ and a learning rate of $0p0005$ For SGD we use a weight decay of $3/5 \times 10^{-4}$ and a learning rate of $0.1$. We plot the validation accuracy curve for CIFAR-$100$ in Figure \ref{subfig:vgg16test}, whilst we see clearly Gadam clearly generalise better than SGD. As shown in Figures \ref{subfig:vgg16sgdbntrain} and \ref{subfig:vgg16gadambntrain}, the spectral norm of the better performing Gadam solutions is almost $40\times$ larger than the SGD solution, the Frobenius norm of Gadam is $0p02$ as opposed to $0p0001$ for SGD. Both solutions give similar training performance, with Gadam $99.81$ and SGD $99.64$. For CIFAR-$10$ although the generalisation gap is smaller, we see a similar picture, as shown in Figure \ref{fig:gadamc10spectrain}. The Frobenius norm of the Gadam solution is $1.55 \times 10^{-3}$ as opposed to $1.43 \times 10^{-5}$.

When batch normalisation is set to evaluation mode, we find a very similar picture. With the spectral norm still $35 \times$ larger for the better performing solution. The Frobenius norm is also $0p04$ instead of $1.3 \times 10^{-5}$ for CIFAR-$100$ and $0p012$ instead of $7.7 \times 10^{-7}$ for CIFAR-$10$.

\begin{figure}[h!]
    \centering
     \begin{subfigure}{0.23\linewidth}
        \includegraphics[width=1\linewidth,trim={0 0 0 0},clip]{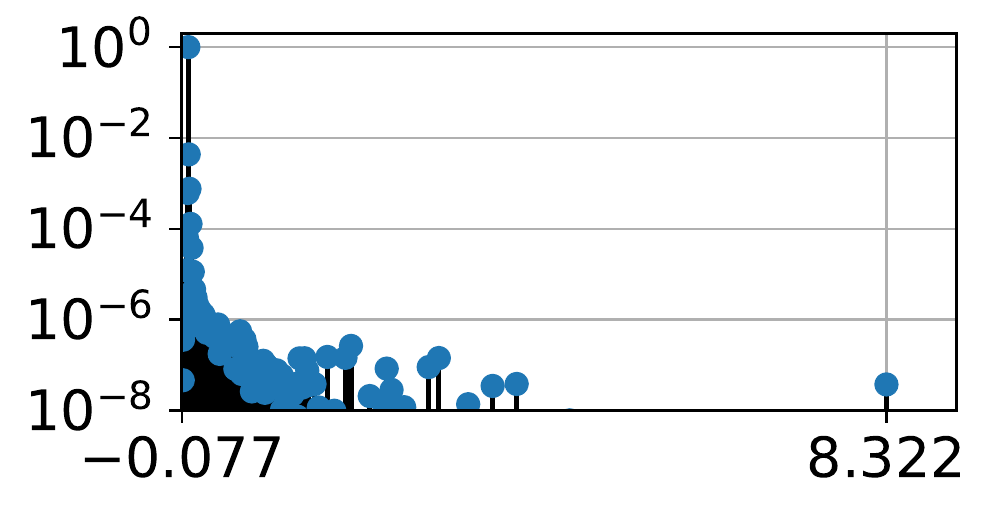}
        \caption{SGD CIAR-$100$}
        \label{subfig:c100sgd}
    \end{subfigure}
    \begin{subfigure}{0.23\linewidth}
        \includegraphics[width=1\linewidth,trim={0 0 0 0},clip]{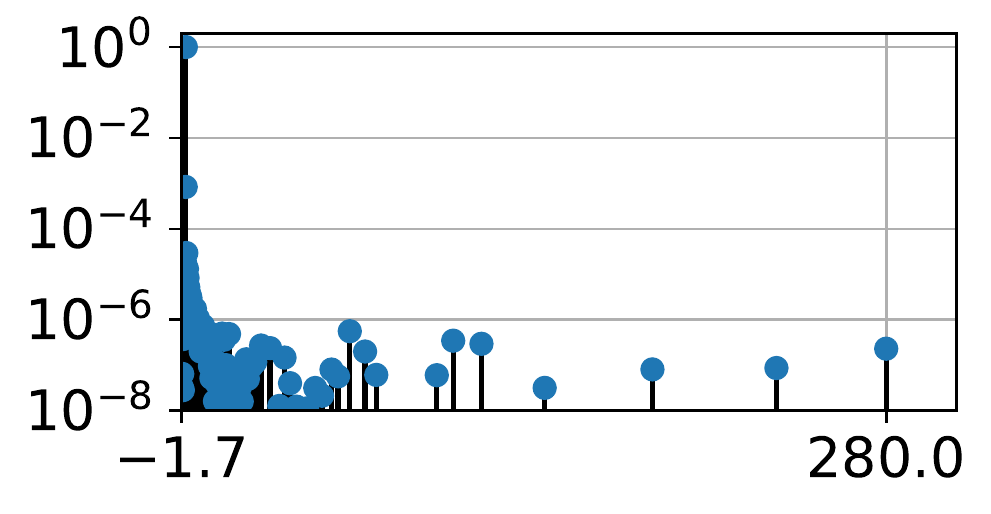}
        \caption{Gadam CIFAR-$100$}
        \label{subfig:c100gadameval}
    \end{subfigure}
     \begin{subfigure}{0.23\linewidth}
        \includegraphics[width=1\linewidth,trim={0 0 0 0},clip]{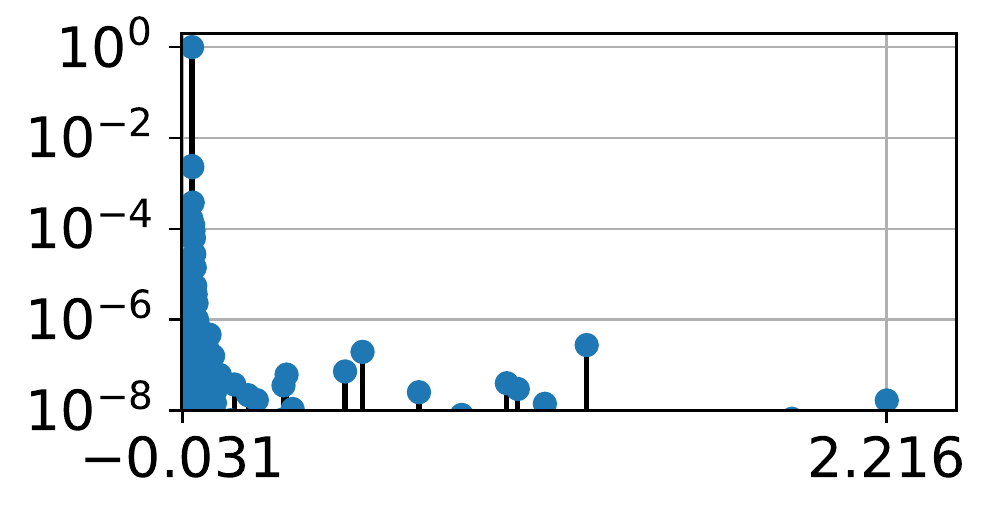}
        \caption{SGD CIFAR-$10$}
        \label{subfig:c10sgdeval}
    \end{subfigure}
    \begin{subfigure}{0.23\linewidth}
        \includegraphics[width=1\linewidth,trim={0 0 0 0},clip]{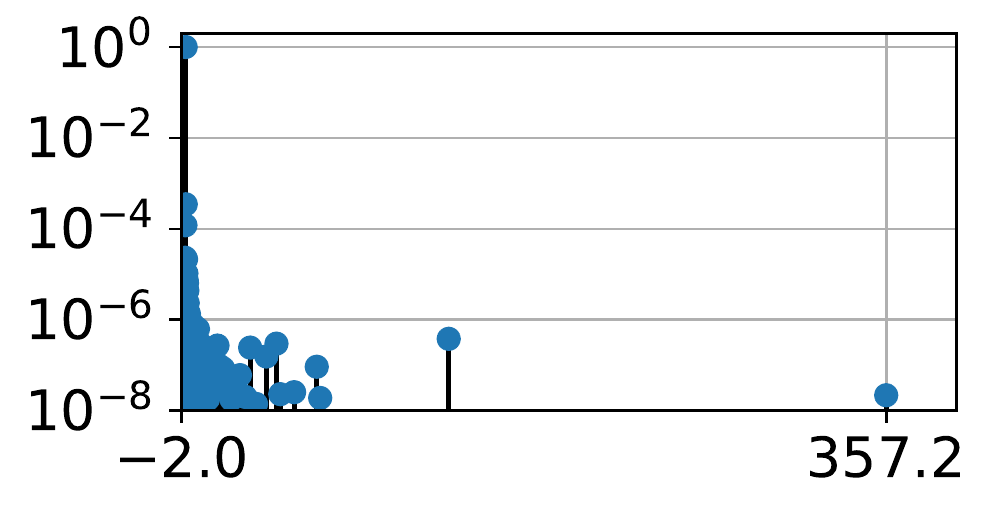}
        \caption{Gadam CIFAR-$10$}
        \label{subfig:c100gadam}
    \end{subfigure}
    \caption{Hessian spectrum for VGG-$16$BN after $300$ epochs of SGD on the CIFAR-$100$ dataset, for various optimisation algorithms [SGD, Gadam], batch norm evaluation mode}
    \label{fig:gadamspeceval}
    
\end{figure}

\section{Conclusion}
In this paper, we consider the deep linear model under the exponential loss, which we analytically show has similar properties to the feed forward neural network under the cross entropy loss. We derive a bound for the rank degeneracy of the Hessian and reason that Hessian based measures of flatness should tend to zero as the loss also tends to zero. Based on this intuition we investigate whether $L2$ regularisation, known to improve generalisation, makes the Hessian sharper. We find that for logistic regression, multi-layer perceptrons, CNNs, pre-activated and wide residual networks, that this is the case. We show that the effect is increased in magnitude when using batch norm in evaluation mode. We further show that alternative adaptive optimisation algorithms which are tuned to give good generalisation performance, can give significantly sharper optima than their non adaptive counterparts. This raises large questions about the applicability of sharp and flat minima in practice as well as theory.
\section{Broader Impact of the work}
Given the importance of achieving strong hold out test set performance for a wide variety of real world applications, understanding what affects generalisation is  a huge and fundamental research question. Certain papers attempt to shed light on this by running a huge amount of experiments, e.g \citep{jiang2019fantastic}. The computational expenditure comes at a huge financial, environmental and opportunity cost. In this paper we give a strong theoretical argument, along with several showcasing experiments as to why traditional Hessian based measures of sharpness, such as the trace, Frobenius and spectral norms should be abandoned when considering generalisation. A potential and very real impact is that algorithms which seek to improve generalisation, or studies which evaluate traditional measures no longer take these measures into account. This could reduce the carbon and financial footprint of this line of work. Since this result is largely theoretical, we do not see immediate impact on democratic institutions, privacy, security or well-being. But related work which may make use of these findings to discover more effective generalisation metrics, could be used to improve models with applications in surveillance eroding privacy or military applications with intent to harm and so we ask that other researchers consider the potential consequences when utilising our work.
\newpage

\bibliography{example_paper}
\bibliographystyle{plainnat}

\appendix

\section{Experiment Details}
\label{sec:experimentdetails}
\subsection{Image Classification Experiments} 

\paragraph{Hyperparameter Tuning} For SGD and Gadam, we set the momentum parameter to be $0.9$ whereas for Adam, we set $(\beta_1, \beta_2) = (0.9, 0.999)$ and $\epsilon = 10^{-8}$, their default values. For SGD, we use a grid searched initial learning rates in the range of  $[0p01, 0p03, 0.1]$ for all experiments with a fixed weight decay; for Adam and all its variants, we use grid searched initial learning rate range of $[10^{-4}, 3 \times 10^{-3}, 10^{-3}]$. After the best learning rate has been identified, we conduct a further search on the weight decay, which we find often leads to a trade off between the convergence speed and final performance. For CIFAR experiments, we search in the range of $[10^{-4}, 10^{-3}]$ whereas for ImageNet experiments, we search in the range of $[10^{-6}, 10^{-5}]$. For decoupled weight decay, we search the same range for the weight decay scaled by initial learning rate.

\subsection{Experimental Details} 
\label{sec:expdetails}
For all experiments with SGD,  we use the following learning rate schedule for the learning rate at the $t$-th epoch, similar to \cite{izmailov2018averaging}:
\begin{equation}
    \alpha_t = 
    \begin{cases}
      \alpha_0, & \text{if}\ \frac{t}{T} \leq 0.5 \\
      \alpha_0[1 - \frac{(1 - r)(\frac{t}{T} - 0.5)}{0.4}] & \text{if } 0.5 < \frac{t}{T} \leq 0.9 \\
      \alpha_0r, & \text{otherwise}
    \end{cases}
\end{equation}
where $\alpha_0$ is the initial learning rate. In the motivating logistic regression experiments on MNIST, we used $T = 50$. $T = 300$ is the total number of epochs budgeted for all CIFAR experiments. We set $r = 0p01$ for all experiments. For experiments with iterate averaging, we use the following learning rate schedule instead:
\begin{equation}
    \alpha_t = 
    \begin{cases}
      \alpha_0, & \text{if}\ \frac{t}{T_{avg}} \leq 0.5 \\
      \alpha_0[1 - \frac{(1 - \frac{\alpha_{avg}}{\alpha_0})(\frac{t}{T} - 0.5)}{0.4}] & \text{if } 0.5 < \frac{t}{T_{avg}} \leq 0.9 \\
      \alpha_{avg}, & \text{otherwise}
    \end{cases}
\end{equation}
where $\alpha_{avg}$ refers to the (constant) learning rate after iterate averaging activation, and in this paper we set $\alpha_{avg} = \frac{1}{2}\alpha_0$. $T_{avg}$ is the epoch after which iterate averaging is activated, and the methods to determine $T_{avg}$ was described in the main text. This schedule allows us to adjust learning rate smoothly in the epochs leading up to iterate averaging activation through a similar linear decay mechanism in the experiments without iterate averaging, as described above.

\section{Lanczos algorithm}
\label{sec:lanczos}
In order to empirically analyse properties of modern neural network spectra with tens of millions of parameters $N = \mathcal{O}(10^{7})$, we use the Lanczos algorithm \citep{meurant2006lanczos}, provided for deep learning by \citet{granziol2019mlrg}. It requires Hessian vector products, for which we use the Pearlmutter trick \citep{pearlmutter1994fast} with computational cost $\mathcal{O}(NP)$, where $N$ is the dataset size and $P$ is the number of parameters. Hence for $m$ steps the total computational complexity including re-orthogonalisation is $\mathcal{O}(NPm)$ and memory cost of $\mathcal{O}(Pm)$. In order to obtain accurate spectral density estimates we re-orthogonalise at every step \citep{meurant2006lanczos}. We exploit the relationship between the Lanczos method and Gaussian quadrature, using random vectors to allow us to learn a discrete approximation of the spectral density.  A quadrature rule is a relation of the form,
	\begin{equation}
		\label{eq:quadraturerule}
		\int_{a}^{b}f(\lambda)d\mu(\lambda) = \sum_{j=1}^{M}\rho_{j}f(t_{j})+R[f]
	\end{equation}
for a function $f$, such that its Riemann-Stieltjes integral and all the moments exist on the measure $d\mu(\lambda)$, on the interval $[a,b]$ and where $R[f]$ denotes the unknown remainder. The nodes $t_{j}$ of the Gauss quadrature rule are given by the Ritz values and the weights (or mass) $\rho_{j}$ by the squares of the first elements of the normalized eigenvectors of the Lanczos tri-diagonal matrix \citep{golub1994matrices}. The main properties of the Lanczos algorithm are summarized in the theorems \ref{theorem:lanczoseigenvalues},\ref{theorem:lanczosspectrum}
\begin{theorem}
	\label{theorem:lanczoseigenvalues}
	Let $H^{N\times N}$ be a symmetric matrix with eigenvalues $\lambda_{1}\geq .. \geq \lambda_{n}$ and corresponding orthonormal eigenvectors $z_{1},..z_{n}$. If $\theta_{1}\geq .. \geq \theta_{m}$ are the eigenvalues of the matrix $T_{m}$ obtained after $m$ Lanczos steps and $q_{1},...q_{k}$ the corresponding Ritz eigenvectors then
	\begin{equation}
	\begin{aligned}
	& \lambda_{1} \geq \theta_{1} \geq \lambda_{1} - \frac{(\lambda_{1}-\lambda_{n})\tan^{2}(\theta_{1})}{(c_{k-1}(1+2\rho_{1}))^{2}} \\
	& \lambda_{n} \leq \theta_{k} \leq \lambda_{m} + \frac{(\lambda_{1}-\lambda_{n})\tan^{2}(\theta_{1})}{(c_{k-1}(1+2\rho_{1}))^{2}} \\
	\end{aligned}	
	\end{equation}
	where $c_{k}$ is the chebyshev polyomial of order $k$
\end{theorem}
Proof: see \citep{golub2012matrix}.
\begin{theorem}
	\label{theorem:lanczosspectrum}
	The eigenvalues of $T_{k}$ are the nodes $t_{j}$ of the Gauss quadrature rule, the weights $w_{j}$ are the squares of the first elements of the normalized eigenvectors of $T_{k}$
\end{theorem}
Proof: See \citep{golub1994matrices}. The first term on the RHS of \eqref{eq:quadraturerule} using Theorem \ref{theorem:lanczosspectrum} can be seen as a discrete approximation to the spectral density matching the first $m$ moments $v^{T}H^{m}v$ \citep{golub1994matrices,golub2012matrix}, where $v$ is the initial seed vector. Using the expectation of quadratic forms, for zero mean, unit variance random vectors, using the linearity of trace and expectation
\begin{equation}
\begin{aligned}
\mathbb{E}_{v}\text{Tr}(v^{T}H^{m}v) & =  \text{Tr}\mathbb{E}_{v}(vv^{T}H^{m}) = \text{Tr}(H^{m})
 = \sum_{i=1}^{N}\lambda_{i} = N \int_{\lambda \in \mathcal{D}} \lambda d\mu(\lambda) \\
\end{aligned}
\end{equation}
The error between the expectation over the set of all zero mean, unit variance vectors $v$ and the monte carlo sum used in practice can be bounded \citep{hutchinson1990stochastic,roosta2015improved}. However in the high dimensional regime $N \rightarrow \infty$, we expect the squared overlap of each random vector with an eigenvector of $H$, $|v^{T}\phi_{i}|^{2} \approx \frac{1}{N} \forall i$, with high probability. This result can be seen by computing the moments of the overlap between Rademacher vectors, containing elements $P(v_{j} = \pm 1) = 0.5$. Further analytical results for Gaussian vectors have been obtained \citep{cai2013distributions}.

\section{Mathematical Preliminaries}

For an input/output pair $[\vx \in \mathbb{R}^{d_{x}},\vy \in \mathbb{R}^{d_{y}}]$ and a given model $h(\cdot;\cdot):\mathbb{R}^{d_{x}}\times \mathbb{R}^{P} \rightarrow \mathbb{R}^{d_{y}}$. Without loss of generality, we consider the family of models functions parameterized by the weight vector $\vw$, i.e., $\mathcal{H}:= \{h(\cdot;\vw):\vw \in \mathbb{R}^{P} \}$, with a given loss $\ell(h(\vx;\vw), \vy): \mathbb{R}^{d_{y}} \times \mathbb{R}^{d_{y}} \rightarrow \mathbb{R}$. 

% The population risk $R_{pop}$ is given as the expectation of the loss over the data-generating distribution $)d\psi(\vx,\vy) $
% \begin{equation}
% 	\label{eq:truerisk}
% 	\mathbb{E}[\ell(h(\vx;\vw),\vy)] = \int_{\mathbb{R}^{d_{x}}\times \mathbb{R}^{d_{y}}}\ell(h(\vx;\vw),\vy)d\psi(\vx,\vy) 
% \end{equation}
% with corresponding gradient $\vg_{pop}(\vw)= \nabla R_{pop}(\vw)$  and Hessian $\mH_{pop}(\vw) = \nabla \nabla R_{true}(\vw)$. 
The empirical risk (often denote the \textit{loss} in deep learning), its gradient and Hessian are given by
\begin{equation}
	\label{eq:emprisk}
	R_{emp}(\vw) = \frac{1}{N}\sum_{i=1}^{N}\ell(h(\vx_{i};\vw),\vy_{i}), \thinspace \vg_{emp}(\vw) = \nabla R_{emp}, \thinspace \mH_{emp}(\vw) = \nabla^{2} R_{emp}
\end{equation}
The Hessian describes the curvature at that point in weight space $\vw$ and hence the risk surface can be studied through the Hessian. By the spectral theorem,  we can rewrite $\mH_{emp}(\vw) = \sum_{i=1}^{P}\lambda_{i}\vphi_{i}\vphi^{T}_{i}$ in terms of its eigenvalue, eigenvector pairs $[\lambda_{i},\vphi_{i}]$. In order to characterise $\mH_{emp}(\vw)$ by a single value, authors typically consider the spectral norm, which is given by the largest eigenvalue of $\mH_{emp}(\vw)$ or the normalised trace, which gives the mean eigenvalue. 
% and represents an upper bound within the quadratic approximation for the increase in loss when perturbing the weight vector. Or the normalised trace, the which gives the mean eigenvalue and the expected change in loss when the weight vector is perturbed by $\dw$.
% \begin{equation}
%     R_{emp}(\vw_{opt}+\dw)-R_{emp}(\vw_{opt}) \approx \frac{1}{2}\dw^{T}\mH(\vw_{opt})\dw = \frac{1}{2}\sum_{i}^{P}\lambda_{i}(\dw^{T}\vphi)^{2} \leq \frac{\lambda_{max}}{2}
% \end{equation}
% Where we have coupled the quadratic approximation near the final weight vector $\vw_{opt}$, gradient at the minimum to be negligible along with the Cauchy-Schwarz inequality. 
The Hessian contains $P^{2}$ elements, so cannot be stored or eigendecomposed for all but the simplest of models. Stochastic Lanczos Quadrature can be used \cite{meurant2006lanczos}, with computational complexity $\mathcal{O}(P)$ to give tight bounds on the extremal eigenvalues and good estimations of $\Tr(\mH)$ and $\Tr(\mH^{2})$, along with a moment matched approximation of the spectrum. We use the Deep Learning implementation provided by \citet{granziol2019mlrg}. DNNs are typically trained using stochastic gradient descent with momentum, where we iteratively update the weights 
\begin{equation}
\label{eq:momentumsgd}
\begin{aligned}
& \vz_{k+1} \leftarrow \rho \vz_{k} + \nabla R(\vw_{k})\\
& \vw_{k+1} \leftarrow \vw_{k} - \alpha \vz_{k+1}
\end{aligned}
\end{equation}
Where $\rho$ is the momentum. The gradient is usually taken on a randomly selected sub-sample of size $B \ll N$. An epoch is defined as a full training pass of the data, so comprises $\approx N/B$ iterations. Often $L2$ regularisation (also termed weight decay) is added to the loss, which corresponds to $R_{emp}(\vw) \rightarrow R_{emp}(\vw)+\mu/2||\vw||^{2}$.%, where $\mu$ is often chosen by grid-search and cross-validation.

\section{Low Rank further investigation}
\label{sec:lowrankapprox}
We provide extensive experimental validation of the low rank nature for both the VGG-$16$ and PreResNet-$110$ on the CIFAR-$100$ datasets in Sections \ref{subsec:rankapproxvgg16} and \ref{subsec:rankapproxp110}.

\begin{figure}[h!]
    \centering
     \begin{subfigure}{0.23\linewidth}
        \includegraphics[width=1\linewidth,trim={0 0 0 0},clip]{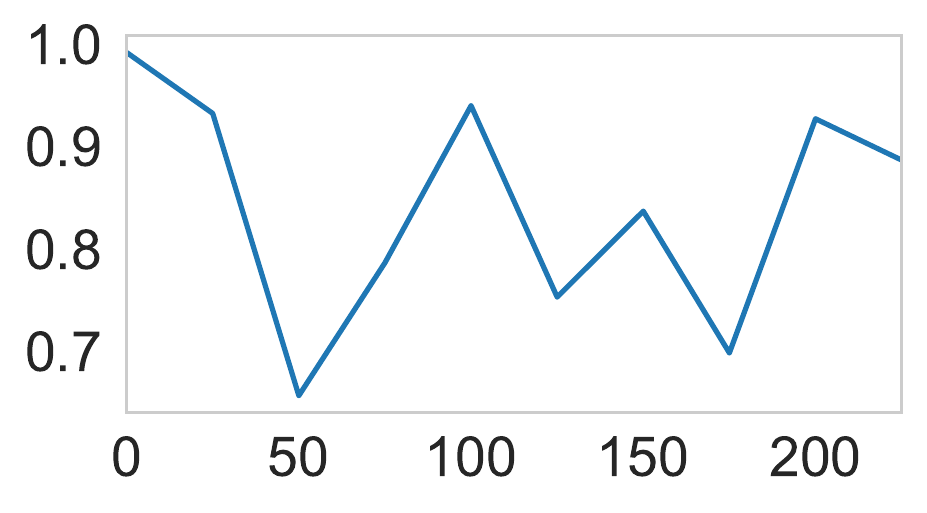}
        \caption{$\mathcal{D}$ weight P110}
        \label{subfig:hessp110prob}
    \end{subfigure}
     \begin{subfigure}{0.23\linewidth}
        \includegraphics[width=1\linewidth,trim={0 0 0 0},clip]{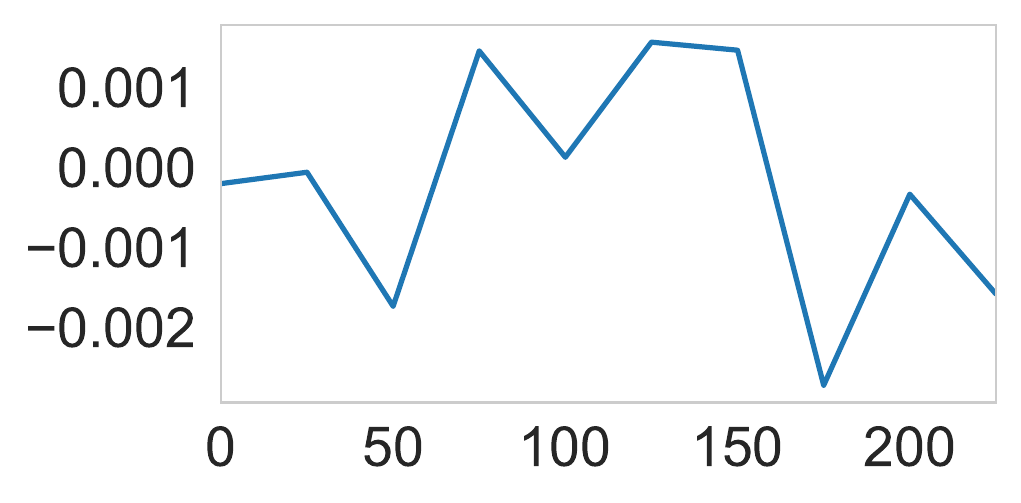}
        \caption{$\mathcal{D}$ value P110}
        \label{subfig:hessp110probv}
    \end{subfigure}
     \begin{subfigure}{0.23\linewidth}
        \includegraphics[width=1\linewidth,trim={0 0 0 0},clip]{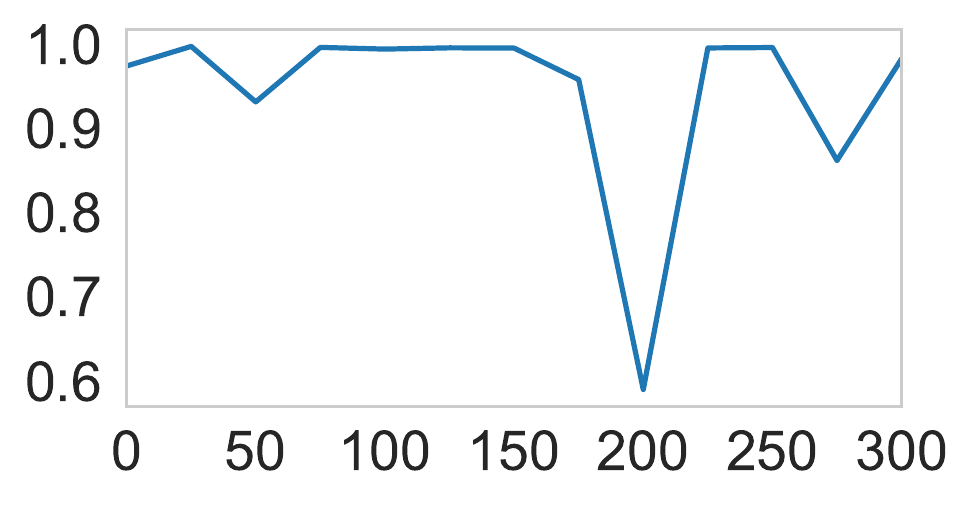}
        \caption{$\mathcal{D}$ weight VGG-$16$}
        \label{subfig:hessvgg16prob}
    \end{subfigure}
     \begin{subfigure}{0.23\linewidth}
        \includegraphics[width=1\linewidth,trim={0 0 0 0},clip]{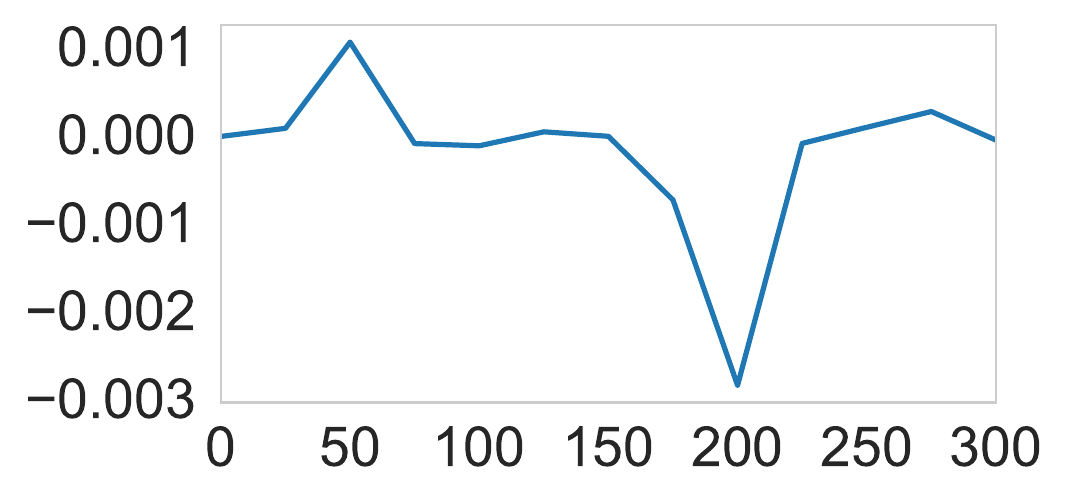}
        \caption{$\mathcal{D}$ value VGG-$16$}
        \label{subfig:hessvgg16probv}
    \end{subfigure}
    \caption{Rank degeneracy $\mathcal{D}$ evolution throughout training using the VGG-$16$ and PreResNet-$110$ on the CIFAR-$100$ dataset, the weight corresponds to the spectral mass of the Ritz value $\mathcal{D}$}
    \label{fig:degenhessprob}
    
\end{figure}

\paragraph{Experimental Setup: }Given that Hessians have $P^{2}$ elements with a full inversion cost of $\mathcal{O}(P^{3})$ which is infeasible for large neural networks. Counting the number of $0$ eigenvalues (which sets the degeneracy) is not feasible in this manner. Furthermore, there would still be issues with numerical precision, so a threshold would be needed for accurate counting. Hence, based on our understanding of the Lanczos algorithm, discussed in section \ref{sec:lanczos}, we propose an alternative method. We know that $m$ steps of the Lanczos method, gives us an $m$-moment matched spectral approximation of the moments of $\vv^{T}\mH\vv$, where in expectation over the set of zero mean unit variance random vectors this is equal to the spectral density of $\mH$. Each eigenvalue, eigenvector pair estimated by the Lanczos algorithm is called a Ritz-value/Ritz-vector. We hence take $m\gg1$, where typically and for consistency we take $m=100$ in our experiments. We then take the Ritz value closest to the origin and take that as a proxy for the $0$ eigenvalue and report its weight. One weakness of this method is that for a large value of $m$, since the Lanczos algorithm finds a discrete moment matched spectral algorithm, is that the spectral mass near the origin, may split into multiple components and counting the largest thereof or closest to the origin may not be sufficient.  We note this problem both for the PreResNet-$110$ and VGG-$16$ on the CIFAR-$100$ dataset shown in Figure \ref{fig:degenhessprob}. Significant drops in degeneracy occur at various points in training and occur in tandem with significant changes in the absolute value of the Ritz value of minimal magnitude. This suggests the aforementioned splitting phenomenon is occurring. This issue is not present in the calculation of the generalised Gauss Newton, as the spectrum is constrained to be positive definite, so there is a limit to the extent of splitting that may occur. In order to remedy this problem, for the Hessian we calculate the combination of the two closest Ritz values around the centre and combine their mass. We consider this mass and the weighted average of their values as the degenerate mass. An alternative approach could be to kernel smooth the Ritz weights at their values, but this would involve another arbitrary hyper-parameter $\sigma$.

\subsection{VGG16}
\label{subsec:rankapproxvgg16}
For the VGG-$16$, which forms the reference model for this paper, we see that for both the generalised Gauss-Newton (shown in Figure \ref{subfig:ggndegenvgg16}) and the Hessian (shown in Figure \ref{subfig:hessdegenvgg16}) that the rank degeneracy is extremely high. For the GGN, the magnitude of the Ritz value which we take to be the origin, is extremely close to the threshold for GPU precision, as shown in Figure \ref{subfig:ggndegenvgg16}. For the Hessian, for which we combine the two smallest absolute value Ritz values, we have as expected an even larger spectral degeneracy. The weighted average, also gives a value very close to $0$, as shown in Figure \ref{subfig:hessdegenvgg16}. Although the combined weighted average is much closer to the origin, than that of the lone spectral peak, shown in Figure \ref{fig:degenhessprob}, which indicates splitting, we do not get as close to the GPU precision threshold.

\begin{figure}[h!]
    \centering
     \begin{subfigure}{0.23\linewidth}
        \includegraphics[width=1\linewidth,trim={0 0 0 0},clip]{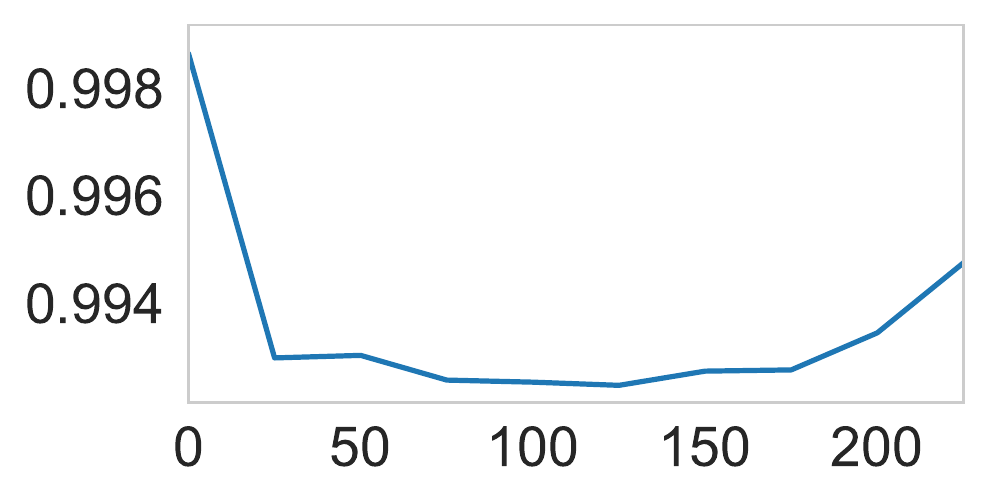}
        \caption{GGN Degeneracy}
        \label{subfig:ggndegenvgg16}
    \end{subfigure}
     \begin{subfigure}{0.23\linewidth}
        \includegraphics[width=1\linewidth,trim={0 0 0 0},clip]{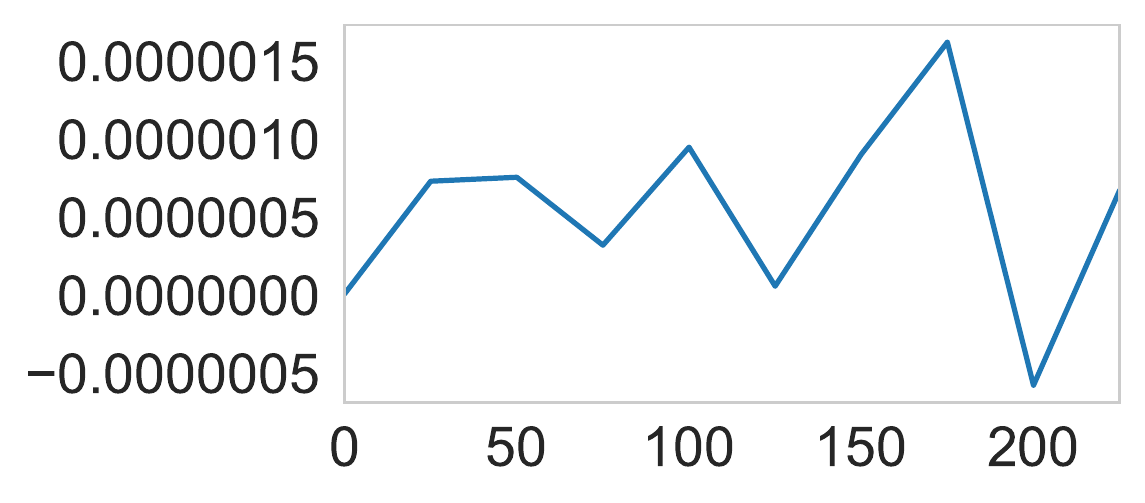}
        \caption{GGN Ritz Value}
        \label{subfig:ggndegenvgg16}
    \end{subfigure}
     \begin{subfigure}{0.23\linewidth}
        \includegraphics[width=1\linewidth,trim={0 0 0 0},clip]{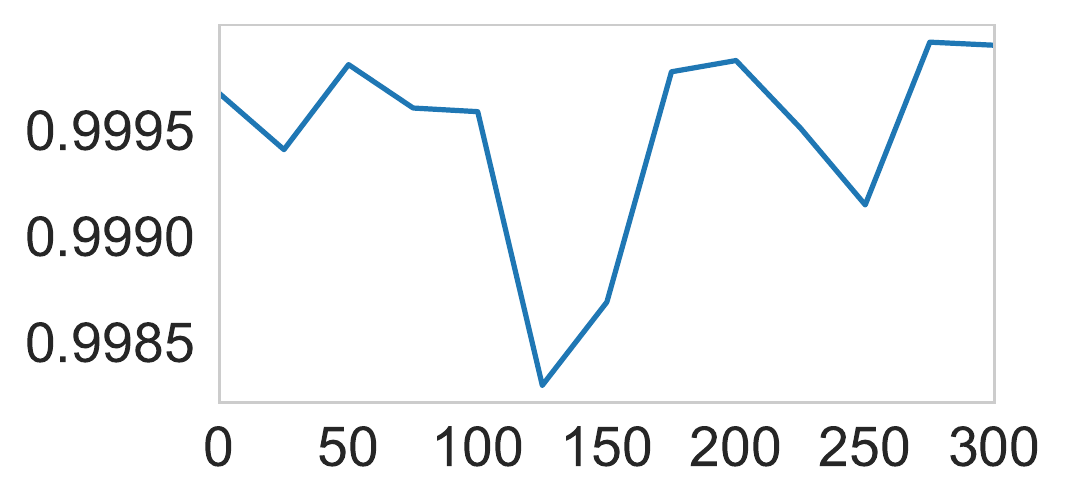}
        \caption{Hessian Degeneracy}
        \label{subfig:hessdegenvgg16}
    \end{subfigure}
     \begin{subfigure}{0.23\linewidth}
        \includegraphics[width=1\linewidth,trim={0 0 0 0},clip]{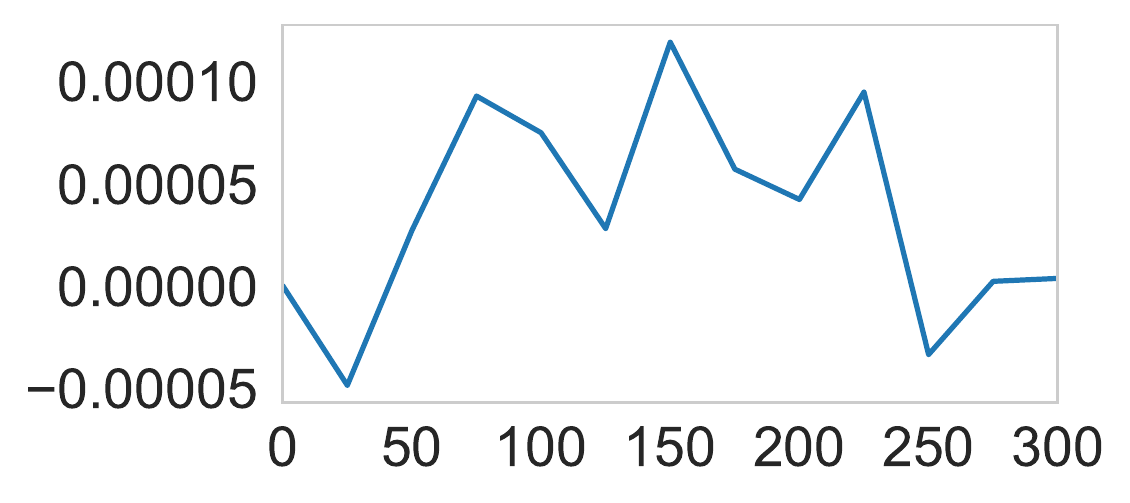}
        \caption{Hessian Ritz Value}
        \label{subfig:hessdegenvgg16}
    \end{subfigure}
    \caption{Rank degeneracy evolution throughout training using the VGG-$16$ on the CIFAR-$100$ dataset, total training $225$ epochs, the Ritz value corresponds to the value of the node which we assign to $0$}
    \label{fig:degenvgg16}
    
\end{figure}
\subsection{PreResNet110}
\label{subsec:rankapproxp110}
We repeat the same experiments in section \ref{subsec:rankapproxvgg16} for the preactivate residual network with $110$ layers, on the same dataset. The slight subtlety is that as explained in Section \ref{sec:batchnormresults}, we can calculate the spectra in both batch normalisation and evaluation mode. Hence we report results for both, with the main finding, that the empirical Hessian spectra are consistent with large rank degeneracy.
\subsubsection{Generalised Gauss Newton}
\begin{figure}[h!]
    \centering
     \begin{subfigure}{0.23\linewidth}
        \includegraphics[width=1\linewidth,trim={0 0 0 0},clip]{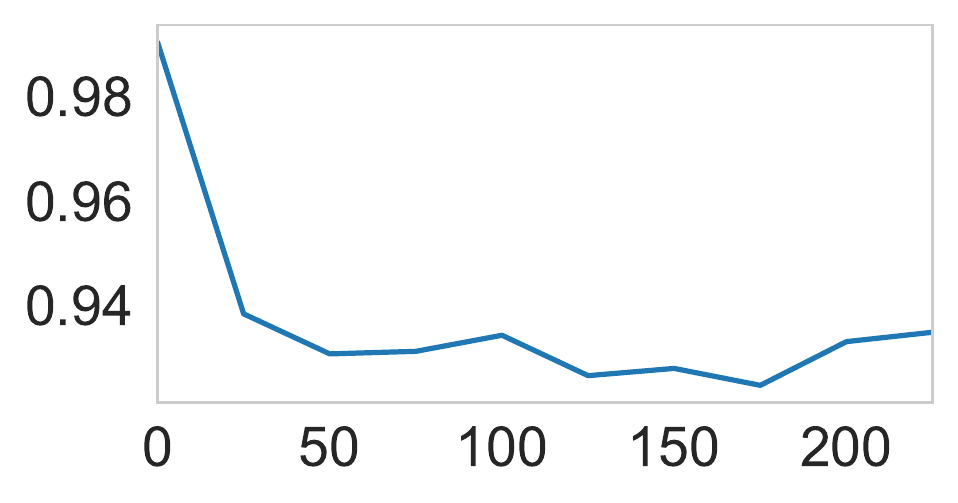}
        \caption{BN-train Degeneracy}
        \label{subfig:ggndegenbnon}
    \end{subfigure}
     \begin{subfigure}{0.23\linewidth}
        \includegraphics[width=1\linewidth,trim={0 0 0 0},clip]{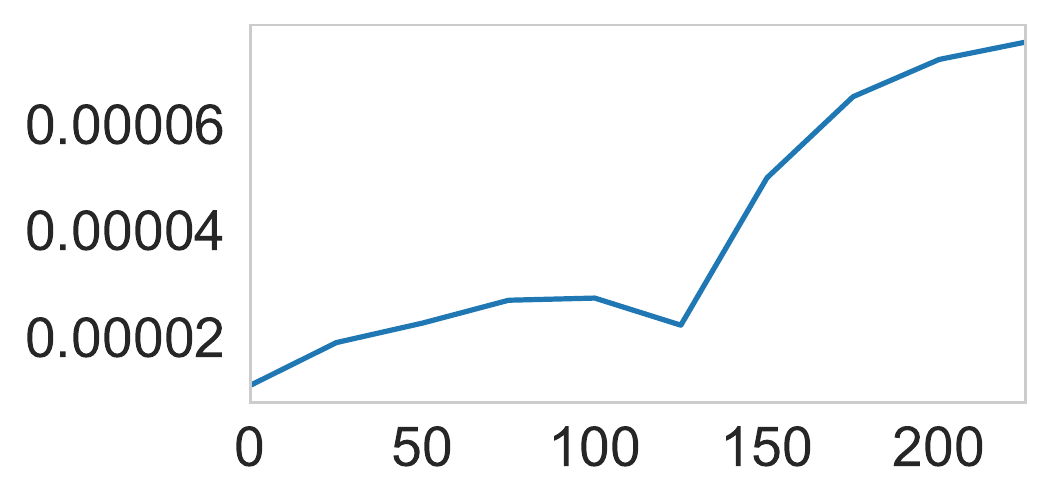}
        \caption{BN-train Ritz Value}
        \label{subfig:ggndegenbnonval}
    \end{subfigure}
     \begin{subfigure}{0.23\linewidth}
        \includegraphics[width=1\linewidth,trim={0 0 0 0},clip]{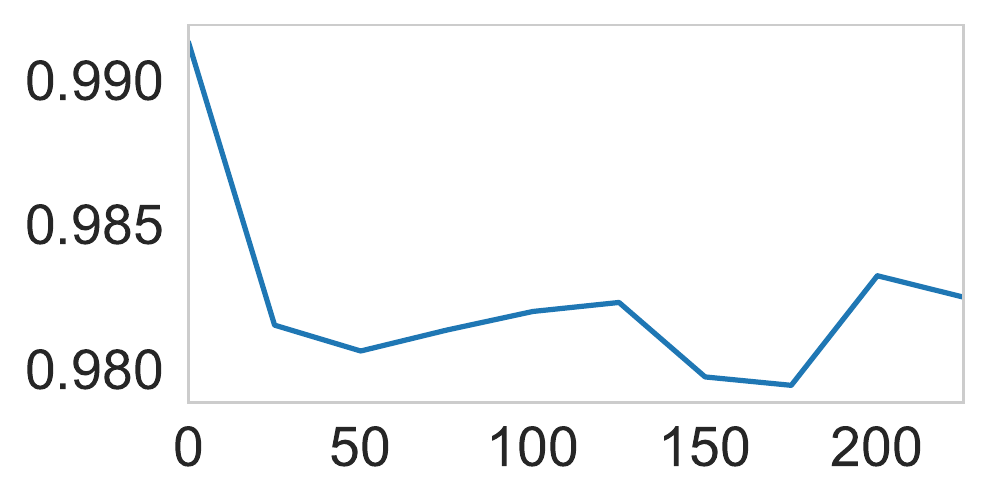}
        \caption{BN-eval Degeneracy}
        \label{subfig:ggndegenbnoff}
    \end{subfigure}
     \begin{subfigure}{0.23\linewidth}
        \includegraphics[width=1\linewidth,trim={0 0 0 0},clip]{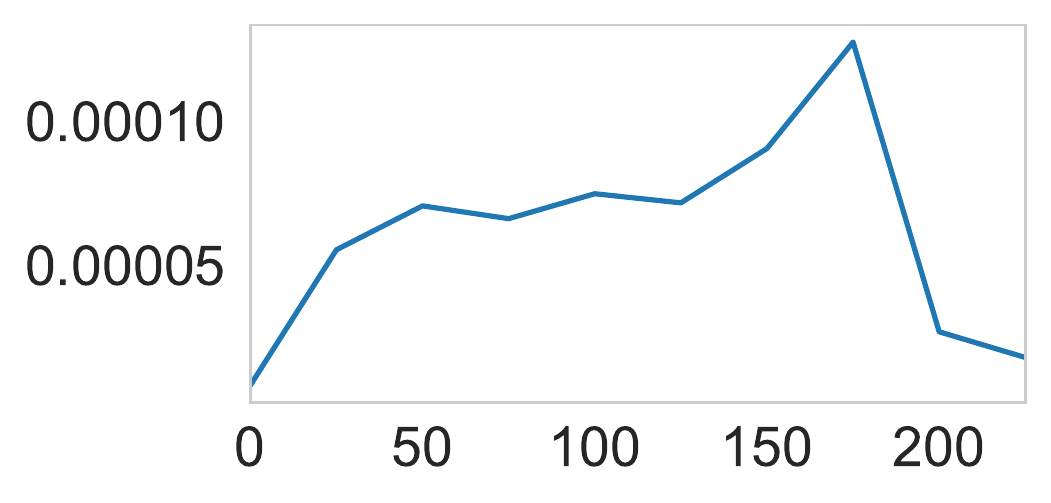}
        \caption{BN-eval Ritz Value}
        \label{subfig:ggndegenbnoffval}
    \end{subfigure}
    \caption{Generalised Gauss Newton rank degeneracy evolution throughout training using the PreResNet-$110$ on the CIFAR-$100$ dataset, total training $225$ epochs, the Ritz value corresponds to the value of the node which we assign to $0$}
    \label{fig:ggndegen}
    
\end{figure}
\subsubsection{Hessian}
\begin{figure}[h!]
    \centering
     \begin{subfigure}{0.23\linewidth}
        \includegraphics[width=1\linewidth,trim={0 0 0 0},clip]{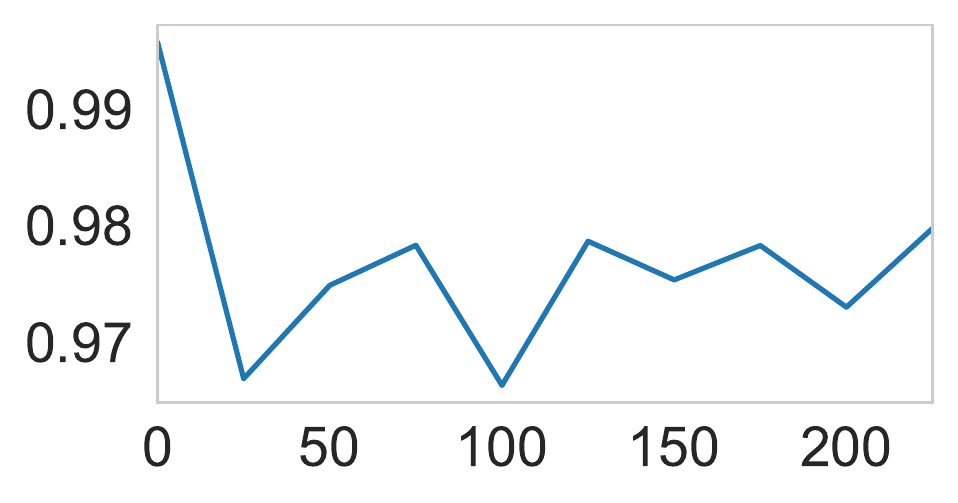}
        \caption{BN-train Degeneracy}
        \label{subfig:hessdegenbnon}
    \end{subfigure}
     \begin{subfigure}{0.23\linewidth}
        \includegraphics[width=1\linewidth,trim={0 0 0 0},clip]{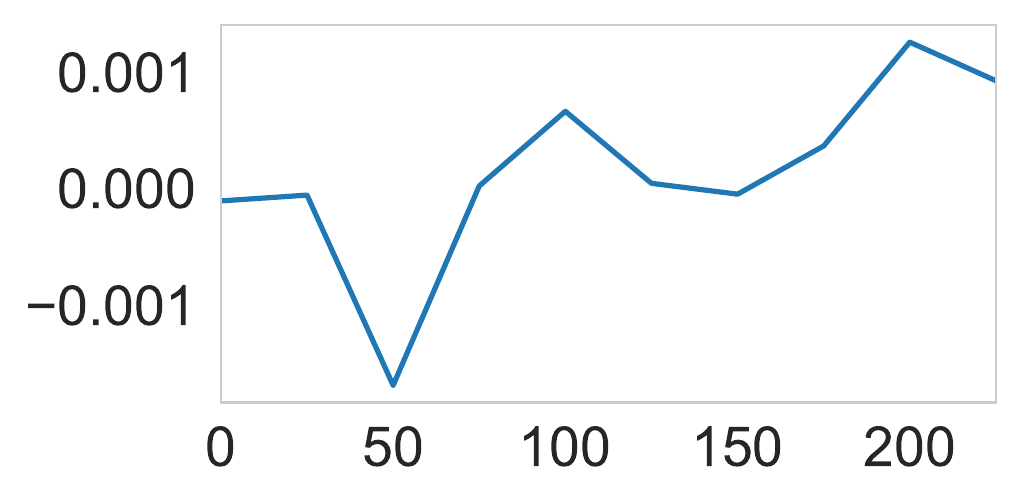}
        \caption{BN-train Ritz Value}
        \label{subfig:hessdegenbnonval}
    \end{subfigure}
     \begin{subfigure}{0.23\linewidth}
        \includegraphics[width=1\linewidth,trim={0 0 0 0},clip]{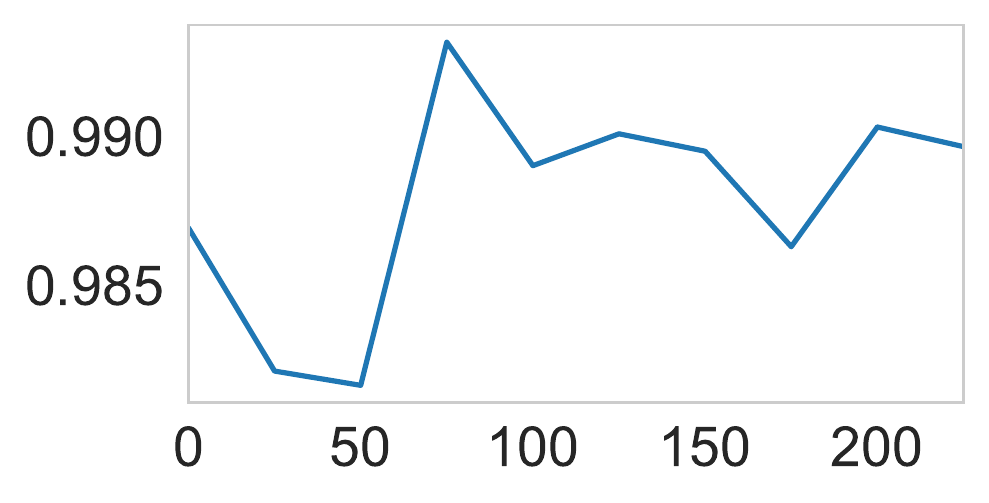}
        \caption{BN-eval Degeneracy}
        \label{subfig:hessdegenbnoff}
    \end{subfigure}
     \begin{subfigure}{0.23\linewidth}
        \includegraphics[width=1\linewidth,trim={0 0 0 0},clip]{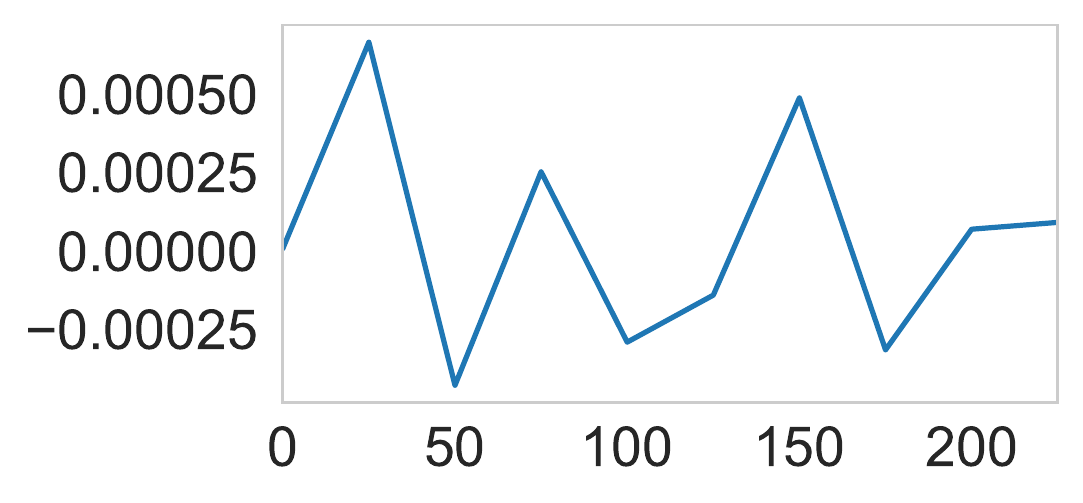}
        \caption{BN-eval Ritz Value}
        \label{subfig:hessdegenbnoffval}
    \end{subfigure}
    \caption{Hessian rank degeneracy evolution throughout training using the PreResNet-$110$ on the CIFAR-$100$ dataset, total training $225$ epochs, the Ritz value corresponds to the value of the node which we assign to $0$}
    \label{fig:hessdegen}
    
\end{figure}

\end{document}